\pdfoutput=1
\def\isarxivbuild{}
\documentclass{article}
\usepackage{amd}
\newcommand{\AMDlogo}{\includegraphics[height=0.28in]{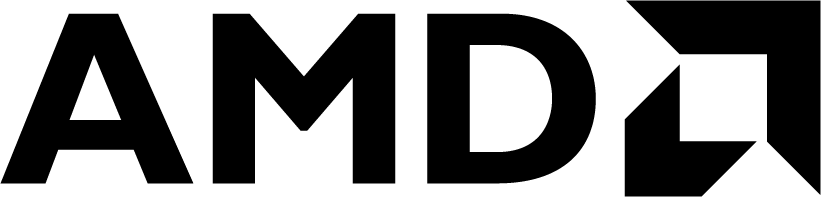}}

\usepackage{hyperref}
\usepackage{microtype}
\usepackage{booktabs}
\usepackage{array}
\usepackage{longtable}
\usepackage{capt-of}
\usepackage{enumitem}
\usepackage{amsmath,amssymb}
\usepackage{tikz}
\usetikzlibrary{arrows.meta,positioning,fit,calc}

\newcommand{\sys}{ROCm InferaSim}

\newcommand{\MIXSLOTS}{23}
\newcommand{\MIXEXACT}{14}
\newcommand{\MIXWTEN}{16}
\newcommand{\MIXBAND}{12\%}

\newcommand{\MIXREG}{0\%}
\newcommand{\MIXCONF}{123}

\newcommand{\ParModels}{13}
\newcommand{\ParPoints}{169,962}
\newcommand{\ParTracePoints}{8,490}

\newcommand{\PMPoints}{529}
\newcommand{\PMCharts}{25}
\newcommand{\PMTputErr}{7.0\%}
\newcommand{\PMIntErr}{6.5\%}
\newcommand{\PMRho}{0.99}
\newcommand{\PMHit}{170}
\newcommand{\PMFront}{206}
\newcommand{\PMHitPct}{83\%}
\newcommand{\PMLossNinety}{0.7\%}
\newcommand{\PMDisagg}{159}

\newcommand{\para}[1]{\smallskip\noindent\textbf{#1}\hspace{0.4em}}
\newif\ifamdattribution \amdattributionfalse
\newcommand{\amdattribution}{\ifamdattribution\par\smallskip\noindent{\footnotesize\copyright{} 2026 Advanced Micro Devices, Inc. All rights reserved. AMD, the AMD Arrow logo, AMD Instinct, Infinity Fabric, ROCm, and combinations thereof are trademarks of Advanced Micro Devices, Inc. Other product names used in this publication are for identification purposes only and may be trademarks of their respective owners.\par}\fi}

\begin{document}
\amdtitlerunning{ROCm InferaSim: Cut LLM Serving Trial Costs and Deploy the Best Configuration}

\amdtitle{ROCm InferaSim: Cut LLM Serving Trial Costs\\ and Deploy the Best Configuration}

\vspace*{-0.5\baselineskip}
{\centering\setlength{\parskip}{0pt}%
  {\bfseries
    Anshu Raina\qquad
    Yao Fu\qquad
    Zhang Jiejing\qquad
    Peyman Razaghi\\[0.35em]
    Shiksha Patel\qquad
    Ravi Gupta\qquad
    Deval Shah\qquad
    Adeem Jassani\\[0.35em]
    Subhajit Dutta Chowdhury\qquad
    Lorri Rao\qquad
    Kailash Gogineni\\[0.35em]
    Zhenyu Gu\qquad
    Onil Gunawardana\qquad
    Emad Barsoum\par}
  \vspace{0.75em}
  Advanced Micro Devices Inc. (AMD)\par
  \vspace{0.6em}
  {\small Correspondence to: \texttt{anshu.raina@amd.com}\par}
}
\amdkeywords{LLM serving, performance modeling, simulation, autotuning}
\vskip 0.15in

\begin{abstract}
Serving a large language model (LLM) efficiently requires making several key decisions: the serving engine, how the model is split across graphics processing units (GPUs), numeric precisions, scheduler settings, speculative decoding, key--value (KV) cache management, and prefill--decode disaggregation. The best configuration depends on the model, hardware, workload, and serving objective (interactivity, time to first token, or throughput). Today, operators typically rely on trial and error to identify the best combination, with each experiment occupying one or more nodes for several minutes. Exhaustively evaluating all candidate configurations in our study would require approximately 25,000 GPU-hours. We propose \sys{}, a framework that predicts LLM serving performance using a few short measurements. Many configuration choices, such as batch size, prompt length, and GPU count, change the amount of work performed by each GPU without changing the underlying GPU kernels. \sys{} benchmarks each distinct set of kernels once, on at most one node, and we call this measurement an \emph{anchor}. Analytical laws carry the anchor to other deployments (a configuration on a given set of GPUs) that run the same kernels. For deployments spanning multiple nodes, \sys{} first reduces the configuration to a single-node equivalent, measures its performance, and then analytically scales the results to the full deployment. A simulator replays realistic multi-turn serving traffic, while an automated tuning agent uses the predicted performance to search for efficient serving configurations. We evaluate on AMD Instinct\texttrademark{} MI355X GPUs with three engines (vLLM, SGLang, and ATOM). Measuring 11 anchors took 9.1 GPU-hours; measuring the 226 fixed-length configurations they cover took 204 GPU-hours. On 172 configurations that pass our measurement filters, time per output token has 8.1\% median error. On a multi-turn coding workload, predictions are within 8\% on throughput. For 159 additional deployments that run prefill and decode on separate GPUs, throughput is within 5.5\%, although no such deployment is used as an anchor. Using predictions alone, among configurations an anchor covers, a short list of the mix planner's five best candidates contains a deployment within 12\% of the best feasible pairings of model and latency target.

\end{abstract}

\section{Introduction}
\label{sec:intro}

\begin{figure}[t]
\centering
\resizebox{\columnwidth}{!}{%
\begin{tikzpicture}[font=\scriptsize,
  gpu/.style={draw, minimum width=11mm, minimum height=4.2mm, inner sep=0pt, align=center, fill=blue!7},
  pool/.style={draw, rounded corners=1pt, minimum width=20mm, minimum height=4.2mm, align=center},
  arr/.style={-{Stealth[length=1.2mm]}}]
\def\row#1#2#3#4#5#6{%
  \node[anchor=east] at (2.3,#1) {#2};
  \node[gpu] at (2.9,#1) {#3}; \node[gpu] at (4.1,#1) {#4};
  \node[gpu] at (5.3,#1) {#5}; \node[gpu] at (6.5,#1) {#6};}
\node[anchor=east, font=\scriptsize\itshape] at (2.3,0.45) {four GPUs:};
\foreach \g [count=\i from 0] in {GPU 0,GPU 1,GPU 2,GPU 3} \node[font=\tiny\itshape] at (2.9+1.2*\i,0.45) {\g};
\row{0}{TP: split every layer}{$\frac14$ of $W$}{$\frac14$ of $W$}{$\frac14$ of $W$}{$\frac14$ of $W$}
\row{-0.5}{EP: place whole experts $E_i$}{$E_{1,2}$}{$E_{3,4}$}{$E_{5,6}$}{$E_{7,8}$}
\row{-1.0}{attention DP: own requests}{req.\ A}{req.\ B}{req.\ C}{req.\ D}
\row{-1.5}{PP: own a block of layers $L_i$}{$L_{1..8}$}{$L_{9..16}$}{$L_{17..24}$}{$L_{25..32}$}

\node[anchor=east] at (2.3,-2.0) {disaggregation:};
\node[pool, fill=orange!12, minimum width=17mm] (pf) at (3.25,-2.0) {prefill pool};
\node[pool, fill=green!10, minimum width=17mm] (dc) at (6.15,-2.0) {decode pool};
\draw[arr, thick] (pf) -- node[below, font=\tiny, yshift=-0.3mm]{KV cache} (dc);
\end{tikzpicture}}
\caption{Forms of parallelism, each on four GPUs. Disaggregation runs prefill and decode in separate pools and sends the KV cache from one to the other. Replicas repeat a layout behind a router.}
\label{fig:par}
\end{figure}
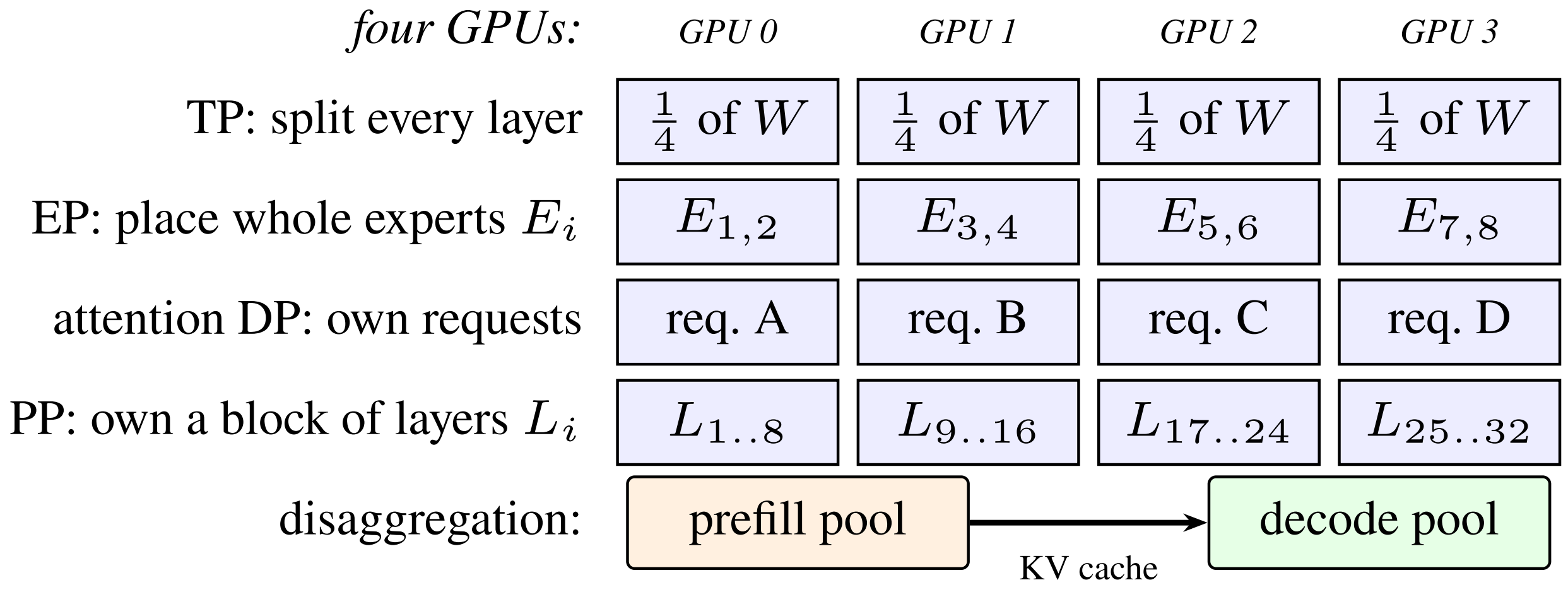

Serving a large language model (LLM) efficiently is a configuration problem. An operator picks a serving engine, such as vLLM~\cite{vllm} or SGLang~\cite{sglang}, and decides how to split the model across graphics processing units (GPUs) (Figure~\ref{fig:par}). Tensor parallelism (TP) splits every layer across GPUs~\cite{megatron}. Expert parallelism (EP) places whole experts of a mixture-of-experts (MoE) model on different GPUs~\cite{gshard}. Data parallelism (DP) gives each GPU its own requests, and engines often apply it to the attention layers only (attention DP)~\cite{deepseekv3,sglang}. Pipeline parallelism (PP) gives each GPU a block of consecutive layers~\cite{gpipe}. TP4 is TP across four GPUs; EP8 and PP2 read alike. The operator also picks precisions for weights, activations and the key--value (KV) cache, scheduler limits, speculative decoding~\cite{specdec}, whether prefill and decode run on separate GPUs (\emph{disaggregated}) or the same ones (\emph{colocated})~\cite{splitwise,distserve}, and the number of replicas. These choices interact, and the best combination changes with the model, the hardware and the traffic. The goal is the cheapest deployment meeting a latency service-level objective (SLO).

\begin{figure}[t]
\centering
\includegraphics[width=\columnwidth]{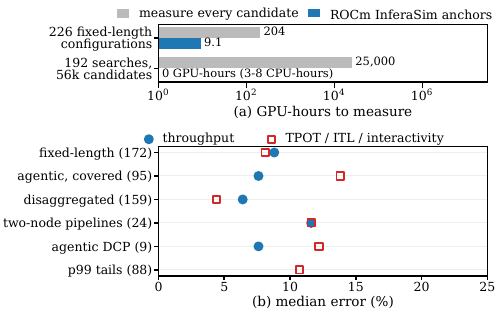}
\caption{\sys{} at a glance. (a) GPU-hours to measure every candidate versus only \sys{}'s anchors. (b) Median prediction error on each validation set (\S\ref{sec:eval}). Agentic sets show the error factor of \S\ref{sec:method} minus one. Tails are 99th-percentile (p99) time per output token (TPOT).}
\label{fig:overview}
\end{figure}

\para{Trial and error is expensive.} Today, each trial loads the model, starts the engine and runs a benchmark. Our measurements show that a configuration holds its GPUs for a median 3.4 minutes, plus engine start-up. Frontier models increasingly require multiple nodes. Evaluating the 226 fixed-length configurations (fixed prompt and output lengths) in our study consumed 204 GPU-hours. Measuring all candidates of the searches in \S\ref{sec:search} would take about 25{,}000 GPU-hours (Figure~\ref{fig:overview}a). Operators therefore tune by hand and deploy the best of a few trials.

\para{Existing tools fall short.} Performance models could eliminate the need for repeated trials, but existing approaches have two main limitations. Operator-level simulators such as Vidur~\cite{vidur} profile each operator and must profile again whenever the model, engine, or kernel library changes. Vidur models only dense layers, and none of these simulators ships data for AMD GPUs (\S\ref{sec:related}). Analytical models~\cite{llmviewer,genz,llmcompass} compute a step's time from hardware peak rates (roofline arithmetic~\cite{roofline}). Although these models capture how computational workloads decrease as more GPUs are added, they overlook the \emph{fixed per-step cost} of kernel launches, synchronization, sampling, and scheduling. This cost remains largely unchanged as the GPU count increases and therefore strongly influences how execution time scales. Ignoring it and naively halving a measured four-GPU DeepSeek-V4-Pro step time to predict eight-GPU performance gives 44\% median error.

\para{Our approach.} \sys{} rests on the key observation that many serving choices change how much work each GPU performs without changing which kernels run. Batch size, context length, TP/EP/PP width, replica count, arrival rate and keeping the KV cache in host memory fall into this category. In contrast, numerical precision, attention implementation, graph-capture mode, speculative decoding method and inference engine can alter the kernels themselves. We call the kernels that one engine step launches its \emph{kernel set}. \sys{} profiles each kernel set once in a short run on at most one node. We call this measurement an \emph{anchor}. Its analytical transfer laws predict other deployments with the same kernel set. They keep the fixed per-step cost of kernel launches, synchronization, sampling and scheduling unchanged while scaling only the GPU-divisible work. On the same four-to-eight-GPU DeepSeek-V4-Pro transfer, this reduces median error to 0.9\% (\S\ref{sec:transfer}).

\para{Beyond one node.} We handle a deployment larger than a node in three steps adapted from training performance prediction~\cite{primusneurips}. We reduce it to a configuration that fits on one node, measure that, and analytically restore what the reduction removed, such as pipeline stages, cross-node communication and wider sharding (\S\ref{sec:rmr}). We also model factors specific to inference: disaggregated pools and their KV-cache transfers, KV caches in host memory, and request queueing.

\para{Agentic traffic and search.} These step predictions describe how a deployment executes, but choosing a deployment also requires modeling the traffic it serves, especially agentic sessions that pause between turns and reuse long prefixes. \sys{} embeds its step model in a discrete-event simulator that captures queueing, batching and prefix reuse. A tuning agent and mix planner then search these predictions instead of evaluating every candidate on GPUs.

\para{Contributions.} This paper makes four contributions:
\begin{itemize}[leftmargin=*,itemsep=0pt,topsep=1pt,parsep=0pt]
\item Anchors and analytical transfer laws that predict every deployment sharing an anchor's kernel set from one measurement (\S\ref{sec:msta}--\S\ref{sec:transfer}).
\item A reduce--measure--restore method for serving that covers pipelines across nodes, disaggregated prefill and decode pools, and KV caches in host memory (\S\ref{sec:rmr}--\S\ref{sec:offload}).
\item A simulate mode that predicts with no measurement at all, used to rank candidates within one engine, and a tuning agent and mix planner that search with the predictor (\S\ref{sec:agent}).
\item An evaluation on AMD Instinct MI355X GPUs with vLLM, SGLang, and the ATOM engine from AMD~\cite{atom}, on fixed-length, agentic and disaggregated traffic (\S\ref{sec:eval}, Figure~\ref{fig:overview}b).
\end{itemize}

\section{Background}
\label{sec:background}

\para{Engines and metrics.} Serving engines process concurrent requests using continuous batching~\cite{orca}, adding and removing requests between steps as they arrive and finish. Each \emph{step} advances every running request by one token and can also process chunks of new prompts, up to a token budget~\cite{sarathi,sarathiserve}. Processing a prompt is \emph{prefill}, and generating output tokens is \emph{decode}. The KV cache is stored in pages~\cite{vllm}, and requests that share a prefix reuse its cached KV~\cite{sglang,mooncake}. Users see time to first token (TTFT), time per output token (TPOT, per request) or inter-token latency (ITL, the gap between consecutive tokens), and throughput. TPOT and ITL have similar means but different tails. TTFT depends mostly on how long a request waits.

\para{Agentic traffic.} Unlike fixed-length benchmarks, which issue independent requests, coding agents generate multi-turn sessions. A turn's prompt can reach hundreds of thousands of tokens, much of it a prefix processed during earlier turns. Agents also pause to ``think'' between turns, creating dependencies in request timing. Agentic workloads therefore combine prefix reuse with interdependent requests. We evaluate both fixed-length and agentic traffic.

\section{Design}
\label{sec:design}

\begin{figure}[t]
\centering
\resizebox{\columnwidth}{!}{%
\begin{tikzpicture}[font=\scriptsize, node distance=3mm and 4mm,
  box/.style={draw, rounded corners=1.5pt, align=center, inner sep=2.5pt, minimum height=5mm},
  arr/.style={-{Stealth[length=1.6mm]}, thick}]
\node[box, fill=gray!10] (spec) {\textbf{1} Serving spec\\{\tiny model $\cdot$ hardware $\cdot$ recipe $\cdot$ workload $\rightarrow$ op graph}};
\node[box, fill=blue!8, below left=4mm and -6mm of spec] (anchor) {\textbf{2} Sparse anchor (\S\ref{sec:anchors})\\{\tiny one step sweep, $\le$1 node}\\{\tiny per kernel set}};
\node[box, fill=orange!12, below right=4mm and -6mm of spec] (sim) {\textbf{3} Kernel models\\{\tiny GEMM, attention, comm., pipeline}\\{\tiny (simulate mode)}};
\node[box, fill=green!10, below=19mm of spec, minimum width=30mm] (kernel) {\textbf{4} Step cost model\\{\tiny step$(b,\text{ctx},\text{TP},\text{EP},\text{PP})$}\\{\tiny + collectives + memory}};
\node[box, below left=5mm and -6mm of kernel] (ana) {\textbf{5} Analytical predictor\\{\tiny means, memory, capacity}};
\node[box, below right=5mm and -6mm of kernel] (des) {\textbf{6} Discrete-event simulator\\{\tiny (\S\ref{sec:des}) sessions, queues, cache}};
\node[box, fill=gray!10, below=17mm of kernel] (search) {\textbf{7} Search (\S\ref{sec:agent}): Pareto sweep / tuning agent\\{\tiny short list $\rightarrow$ confirm on hardware}};
\draw[arr] (spec) -- node[left, font=\tiny, xshift=-0.5mm]{reduce (\S\ref{sec:rmr})} (anchor); \draw[arr] (spec) -- (sim);
\draw[arr] (anchor) -- node[left, font=\tiny, align=right, xshift=-2.5mm, yshift=-1mm]{transfer (\S\ref{sec:transfer}),\\restore (\S\ref{sec:rmr})} (kernel);
\draw[arr] (sim) -- (kernel);
\draw[arr] (kernel) -- (ana); \draw[arr] (kernel) -- (des);
\draw[arr] (ana) -- (search); \draw[arr] (des) -- (search);
\end{tikzpicture}}
\caption{\sys{} architecture. The step cost model (4) is calibrated either by an anchor (2) measured on a target reduced to at most one node (anchored mode) or by kernel models alone (3, simulate mode). The analytical predictor (5) and the discrete-event simulator (6) both use it, and both feed the search (7).}
\label{fig:arch}
\end{figure}
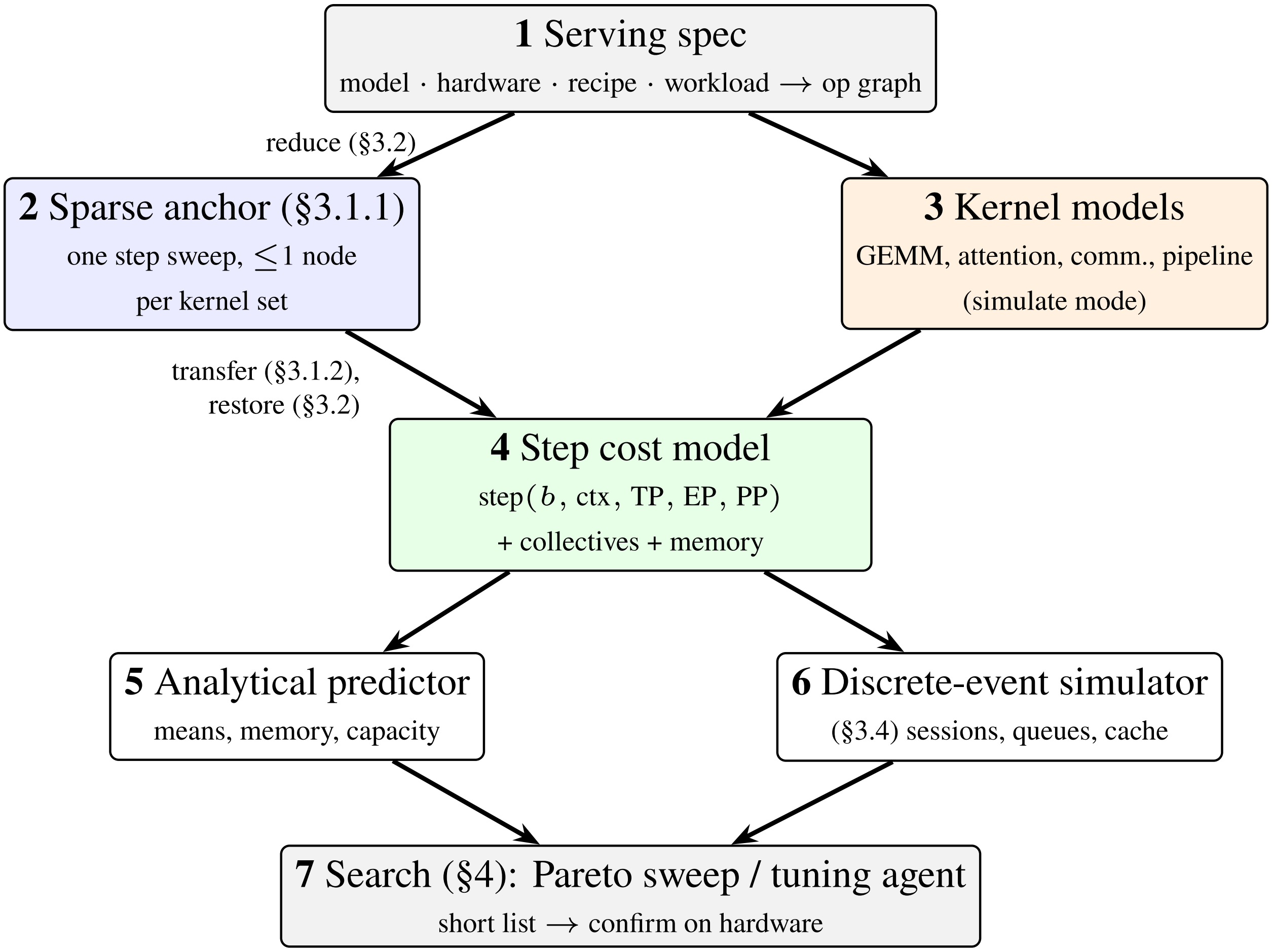

Figure~\ref{fig:arch} shows the structure of \sys{}. At its center is a \emph{step cost model} that predicts the time of one prefill, decode or mixed step for any batch, context length and parallel layout. It runs in one of two modes. In \textbf{anchored mode} it is calibrated by an anchor (\S\ref{sec:anchors}). In \textbf{simulate mode}, \sys{} predicts step time without a GPU measurement. It uses the serving specification and the engine's model code to reconstruct the operations executed in one step. It predicts general matrix multiplication (GEMM) times with Origami~\cite{origami}, and our own analytical models price attention, the other operators, communication (\S\ref{sec:coll}) and pipeline stages (\S\ref{sec:ppnodes}), with a minimum cost per kernel launch. Both modes use the same step model. Anchored mode uses the measured anchor step time as its baseline and applies analytical models to predict how that time changes for widths, batch sizes or context lengths not covered by the anchor. Simulate mode has no measured baseline, so its predictions are used only to rank candidates (\S\ref{sec:rank}).

\subsection{Measure sparsely, transfer analytically}
\label{sec:msta}

\sys{} sorts serving choices into two kinds. A choice is \emph{anchor-defining} if it changes the kernels a step runs, and each combination of such choices needs its own anchor. A choice is \emph{transferable} if it changes only how much work each GPU does, and is predicted from an existing anchor.

\para{Anchor-defining choices.} These are the model checkpoint, the precision of weights, activations, KV cache and experts, the attention and MoE implementations, graph capture, the speculative method, the engine and the GPU type.

\para{Transferable choices.} These are batch size and concurrency, prompt and context length, the TP, EP, PP and attention-DP widths, the number of replicas, the arrival pattern, and KV caching in host memory. The laws of \S\ref{sec:transfer} keep the fixed per-step cost and scale the rest.

\subsubsection{Anchors}
\label{sec:anchors}

An anchor times complete engine steps for one kernel set, and every configuration with that kernel set is predicted from it. We call the measurement run a \emph{harvest}, and the \sys{} tool that performs it the \emph{harvester}. The harvester starts the engine with the anchor's configuration on idle GPUs, runs the sweeps below and writes the results to a JavaScript Object Notation (JSON) file (Appendix~\ref{app:anchor}, uploaded separately). The file holds:
\begin{itemize}[leftmargin=*,itemsep=0pt,topsep=1pt,parsep=0pt]
\item a decode sweep with the median decode step time at each batch size the engine captures as a graph;
\item a prefill sweep over five prompt lengths, fitted as $T(n)=F+c_1n+c_2n^2$ for $n$ prompt tokens;
\item an optional probe of several prompts processed in one step;
\item a \emph{signature} of what transfer relies on: engine version, prompt length, whether real weights were loaded, whether prefix caching was off, and the captured batch sizes.
\end{itemize}

\para{When an anchor applies.} An anchor is used for a target only if all its anchor-defining fields match the target's. \sys{} checks these fields one by one. It does not verify from a kernel trace that the target runs the same kernels. It refuses an anchor from another GPU type or engine and flags one used outside its measured range (Appendix~\ref{sec:lessons}). Anchors are cached by kernel set and TP width, and each pair is measured once. This paper uses 75 cached anchors.

\subsubsection{Transfer laws}
\label{sec:transfer}

\para{Decode across TP width.} Let $s_w(b)$ be the measured decode step time at TP width $w$ and batch $b$. Weight reads and matrix multiplications are split across more GPUs as TP grows, so that part of the step shrinks. The fixed per-step cost does not. Dividing the whole step by the width ratio, $s_{w'}=s_w\,w/w'$, wrongly shrinks both parts. \sys{} instead keeps the fixed part $f$, taken from the anchor's own batch sweep, and scales only the remainder:
\begin{equation}
s_{w'}(b) = f + \bigl(s_w(b)-f\bigr)\cdot \rho_{w\to w'}(b).
\label{eq:floor}
\end{equation}
Here $f$ is the cheapest step of the anchor's decode sweep (its batch-1 step), and $\rho$ is how much the remainder shrinks from width $w$ to $w'$. By default $\rho=w/w'$. Where the same setup was also harvested at half the anchor's width, $\rho$ is measured instead: the per-doubling shrink of $s(b)-s(1)$ between the two widths, at each $b>1$ whose excess is at least 1\,ms, clamped to $[0.5,1]$ and raised to the number of doublings (Appendix~\ref{app:shard}). As a check, we moved a DeepSeek-V4-Pro TP4 anchor to TP8 and compared with measured TP8 steps at batches 1--64. Eq.~\ref{eq:floor} has 0.9\% median error. Two simpler transfers do worse (Table~\ref{tab:component}). Halving the TP4 step has 44.1\% error. Subtracting simulate mode's predicted TP4-to-TP8 drop from the measured TP4 step has 18.5\%, because simulate mode shards the fixed cost too.

\para{Prefill across TP width.} Prefill follows the same idea. The fixed term $F$ is kept and the per-token terms $c_1$ and $c_2$ are scaled with width. For DeepSeek-V4-Pro, solving the TP4 and TP8 prefill steps for $F$ gives 146.9\,ms at 4k-token prompts and 145.9\,ms at 8k, within 1\%.

\para{Decode is a staircase under graph capture.} The engine pads a batch up to the next captured graph size, so a batch of 33 costs as much as 40. Piecewise graphs exempt only attention, so steps soften but remain. \sys{} reads the measured step at the padded size instead of interpolating, which would make in-between batches look cheaper. Eager mode and batches above the largest graph are not padded.

\para{Context length.} A decode step reads the KV cache of every running request and slows down as contexts grow. Anchors measured at two prompt lengths give the cost per extra token of context. Otherwise \sys{} uses the anchor's context and reports the mismatch.

\para{Expert imbalance.} With EP, an MoE layer finishes only when its busiest GPU finishes. Load is uneven by chance at small batch, and because a trained router prefers some experts. \sys{} records the router's choices once per checkpoint and replays them for any EP layout, with or without expert-parallel load balancing (EPLB)~\cite{eplb}, which replicates hot experts. The anchor often has a smaller EP than the target, so \sys{} computes imbalance at each EP. A single value would cancel in the ratio and make wider EP look free (\S\ref{sec:moe-eval}). Simulate mode applies the same replay at the target EP, but we validate imbalance only in anchored mode. Without a recording, it samples a uniform router.

\para{Other components.} Communication, pipeline hops, host time and KV transfers are priced from constants measured once per platform (Table~\ref{tab:constants}, Appendix~\ref{app:models}). Memory use is computed analytically and limits concurrency.

\subsection{Reduce, measure, restore}
\label{sec:rmr}

\begin{table*}[t]
\caption{Reduce, measure, restore. Each row is one way a target can be larger than its anchor. Columns give what the measured run keeps, how \sys{} adds the rest back, and the measured deployments (never anchors) that check it.}
\label{tab:rmr}
\centering\scriptsize\setlength{\tabcolsep}{3pt}
\begin{tabular}{@{}>{\raggedright\arraybackslash}p{0.16\textwidth}>{\raggedright\arraybackslash}p{0.20\textwidth}>{\raggedright\arraybackslash}p{0.30\textwidth}>{\raggedright\arraybackslash}p{0.30\textwidth}@{}}
\toprule
Target exceeds the anchor in & Reduced to & Restored by & Validated on \\
\midrule
TP width & TP/2 or narrower, same node & fixed cost kept, rest scaled (Eq.~\ref{eq:floor}) & 90 TP8 deployments from TP4 anchors: 9.7\% throughput error (\S\ref{sec:fixed}) \\
EP width & the anchor's EP (often 1) & busiest-GPU imbalance at both EPs; the engine's dispatch (App.~\ref{sec:moe}) & nine vLLM EP runs: 2.2\% TPOT; three SGLang EP runs: within 2.5\% \\
Pipeline stages, across nodes & PP1 on one node & a network hop per stage, plus the host time PP exposes (Eq.~\ref{eq:pp}) & 24 two-node pipelines of 2--16 GPUs: 11.6\% TPOT (\S\ref{sec:ppnodes}) \\
Communication across nodes & none in the anchor & $\alpha+m/B$ on in-node and network links (\S\ref{sec:coll}) & 35 two-node collectives: within about 10\% \\
Disaggregated pools & one replica, prefill and decode together & one simulated pool per phase, each with its own layout, plus the KV transfer (\S\ref{sec:disagg}) & 159 splits of 8--24 GPUs: 5.5\% throughput, 5.3\% interactivity \\
Host KV tier (offload) & GPU memory only & a host-memory level in the simulated cache, reloads over the host link (\S\ref{sec:offload}) & 20 covered agentic offload runs (\S\ref{sec:variable}) \\
Load and traffic & a few batch sizes or one load & simulated sessions, queues, prefix cache and replicas (\S\ref{sec:des}) & 172 fixed-length and 95 agentic deployments \\
\bottomrule
\end{tabular}
\end{table*}

Some targets cannot run as they are on one node, because they are wider than the anchor, span nodes, split prefill and decode, or use host memory. For these \sys{} uses three steps adapted from training performance prediction~\cite{primusneurips}. Figure~\ref{fig:arch} places reduction before the anchor (2) and restoration in the step model (4); Table~\ref{tab:rmr} lists cases.
\begin{itemize}[leftmargin=*,itemsep=0pt,topsep=1pt,parsep=0pt]
\item \emph{Reduce} the target to a run with the same kernels, using one replica with prefill and decode together, on at most one node, at one load or a few batch sizes. For a target that spans nodes, \sys{} removes pipeline stages first and then narrows EP and TP until the run holds one stage's kernels.
\item \emph{Measure} that run once to obtain an anchor (\S\ref{sec:anchors}).
\item \emph{Restore} the removed parts (communication, memory, scheduling) with analytical laws and measured constants.
\end{itemize}
Only the restore step depends on the target. One measurement serves all targets with the same kernels, and a 24-GPU disaggregated deployment costs no more to predict than one GPU. We rederive the training method's TP, EP and pipeline laws for prefill and decode steps. Disaggregated pools, host-memory KV and queueing are new.

\subsection{The configuration space}
\label{sec:space}

A deployment combines the parallelism forms of Figure~\ref{fig:par} with precisions, kernel implementations, scheduler settings, speculation and KV memory tiers. Table~\ref{tab:space} (Appendix~\ref{app:design}) lists the choices and how each is priced. Each choice has its own cost term, and two layouts that differ in one choice get different predictions. Illegal combinations are removed (for example, TP must divide the number of attention heads), as are those that do not fit in GPU memory. The Pareto sweep enumerates the space for each model, and the agent searches spaces too large to enumerate (\S\ref{sec:agent}).

\subsection{Discrete-event simulator (DES)}
\label{sec:des}

\para{Why simulate requests.} SLOs are usually set on tail latency, and tails have their own causes. In one measured run with 64 clients, the 99.9th-percentile ITL was 1392\,ms while the median was 21.7\,ms, because each newly admitted prompt stalls all running decodes. A model of averages cannot show this. The DES simulates individual requests and produces latency distributions.

\para{What it simulates.} The DES advances a virtual clock through request arrivals, admission, steps and completions, and the step cost model prices each step. It mimics the target engine's scheduler (token budget, chunked prefill, mixed steps and, for ATOM, separate prefill-only steps), a paged KV cache, a block-level prefix cache that evicts like the engines do, and routing across replicas. The cache hit rate is a result of the simulation, not an input. This scheduler model and the harvester's launch code are the only engine-specific parts of \sys{}. Like our measurements, it runs $C$ clients that each send a new request when the previous one finishes, or open-loop arrivals.

\para{Agentic sessions.} A session's next turn is sent only after the previous turn finishes and its think time passes. Its prefix hit comes from a simulated cache replayed in the engine's admission order (Appendix~\ref{app:des}).

\para{Host time.} Central processing unit (CPU) work per request (tokenization, detokenization, preparing uncached prompts) overlaps the GPU and is added only to TTFT and end-to-end latency. It is fitted per engine from the anchor runs. TTFT is reported both from admission, as engine benchmarks do, and from arrival, as users see it.

\para{Scheduler settings.}\label{sec:knobs} Configurations that differ only in a scheduler setting run the same kernels, and the anchor alone cannot tell them apart. \sys{} models their effect on scheduling in the DES instead. These are the cap on running requests, the chunked-prefill budget, the attention-DP split of requests, the largest captured batch, and the KV capacity a KV-cache precision gives. This separates configurations that would otherwise tie (\S\ref{sec:rank}).

\subsection{KV offload engines}
\label{sec:offload}

A multi-turn trace reuses long prefixes, but GPU high-bandwidth memory (HBM) holds the KV cache of only the most recent sessions. Offload engines add a second, larger tier in host dynamic random-access memory (DRAM): SGLang's HiCache, LMCache~\cite{lmcache} (used by our ATOM runs), vLLM's CPU offload connector and, for disaggregated SGLang, the unified memory and bandwidth pool (UMBP) of MoRI (Modular RDMA Interface)~\cite{mori}, a remote direct memory access (RDMA) library from AMD. Nearly two in five runs of our agentic data use one. Offloading changes which prompt tokens must be recomputed. It leaves the kernels alone and is transferable. The DES models it as a second cache level. Blocks evicted from HBM move to DRAM, and a DRAM hit is copied back before the request runs (Appendix~\ref{sec:kv-tier}).

\subsection{Speculative decoding without an anchor}
\label{sec:spec-sim}

Speculative decoding lets a small drafter propose several tokens that the main model then checks in one step. The drafters we measure are the models' own multi-token-prediction (MTP) heads~\cite{mtp}. An anchor measured with speculation already includes the drafter's cost and acceptance. Otherwise \sys{} prices each drafter layer like a main-model layer of the same kind, and takes the number of accepted tokens at each draft depth from InferenceX~\cite{inferencex}, a public collection of measured serving runs, optionally adjusted per domain with SPEED-Bench~\cite{speedbench}. It picks the draft depth with the most tokens per unit of step time (Appendix~\ref{app:spec}).

\section{The Tuning Agent}
\label{sec:agent}
\label{sec:search-design}

A prediction takes under half a second and orders candidates well (\S\ref{sec:rank}), which makes it usable in place of trials inside a search. The \sys{} tuning agent searches the serving space this way (Figure~\ref{fig:agentflow}, details in Appendix~\ref{app:agent}), following a two-phase design from automated tuning of distributed training. A \emph{lever} is one setting the agent may change, together with its allowed values. There are 36 levers, covering the choices of Table~\ref{tab:space}, such as parallel widths, disaggregation and its pools, precisions, kernel backends, graph capture, scheduler budgets, draft depth, communication algorithms and kernel fusions. The agent optimizes one of 24 predicted objectives, such as throughput, latency, tokens/s per user, capacity or cost.

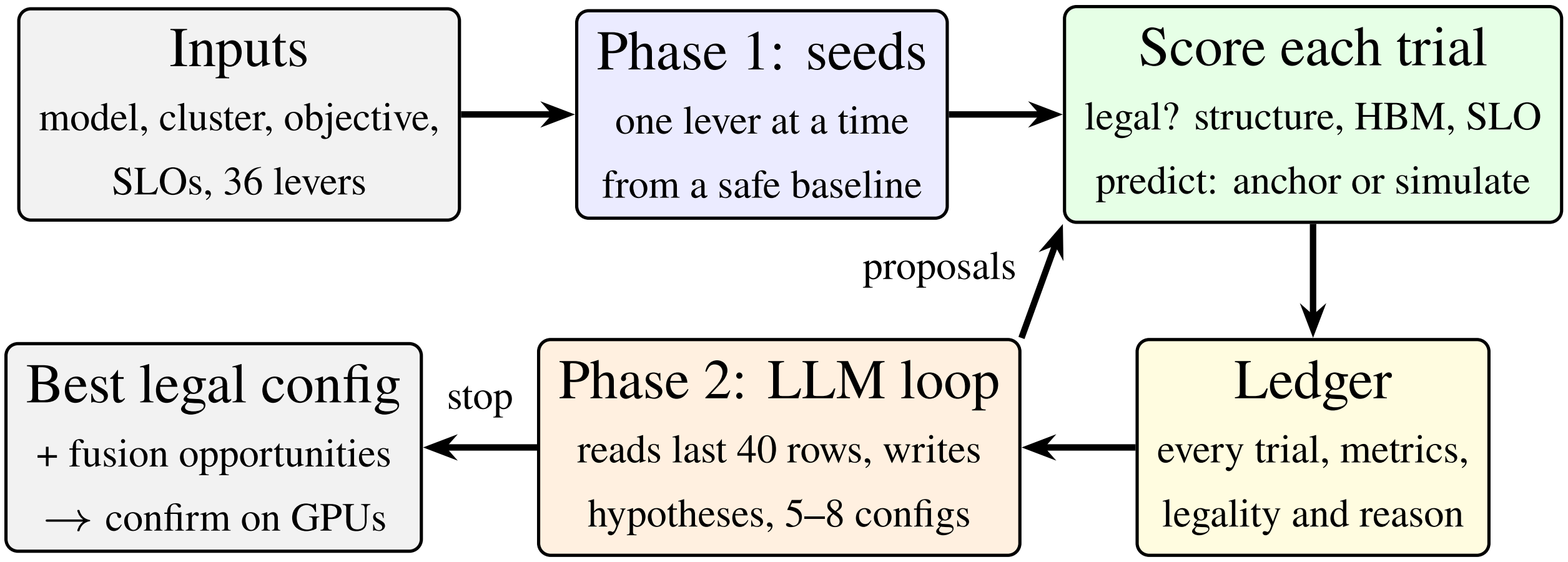
\begin{figure}[t]
\centering
\resizebox{\columnwidth}{!}{%
\begin{tikzpicture}[font=\scriptsize, node distance=4mm and 5mm,
  box/.style={draw, rounded corners=1.5pt, align=center, inner sep=2.5pt, minimum height=8mm},
  arr/.style={-{Stealth[length=1.5mm]}, thick}]
\node[box, fill=gray!10] (in) {Inputs\\{\tiny model, cluster, objective,}\\{\tiny SLOs, 36 levers}};
\node[box, fill=blue!8, right=of in] (seed) {Phase 1: seeds\\{\tiny one lever at a time}\\{\tiny from a safe baseline}};
\node[box, fill=green!10, right=of seed] (score) {Score each trial\\{\tiny legal? structure, HBM, SLO}\\{\tiny predict: anchor or simulate}};
\node[box, fill=yellow!15, below=5mm of score] (led) {Ledger\\{\tiny every trial, metrics,}\\{\tiny legality and reason}};
\node[box, fill=orange!12, left=of led] (llm) {Phase 2: LLM loop\\{\tiny reads last 40 rows, writes}\\{\tiny hypotheses, 5--8 configs}};
\node[box, fill=gray!10, left=of llm] (out) {Best legal config\\{\tiny + fusion opportunities}\\{\tiny $\rightarrow$ confirm on GPUs}};
\draw[arr] (in) -- (seed); \draw[arr] (seed) -- (score);
\draw[arr] (score) -- (led); \draw[arr] (led) -- (llm);
\draw[arr] (llm) -- node[above, font=\tiny]{stop} (out);
\draw[arr] (llm.north east) -- node[left, font=\tiny, pos=0.6]{proposals} (score.south west);
\end{tikzpicture}}
\caption{The tuning agent. Seed trials and LLM proposals go through the same legality check and prediction. The LLM never estimates performance itself. Each trial is one prediction of under half a second, with no GPU.}
\label{fig:agentflow}
\end{figure}

\para{Two phases.} First, a fixed seed planner changes one lever at a time from a safe baseline. Each gain then has a clear cause. Then an LLM loop, written in DSPy~\cite{dspy} in the style of ReAct~\cite{react}, reads the latest 40 trials in the \emph{ledger} (the trial log). It writes hypotheses and 5--8 partial configurations and evaluates them one at a time, combining levers the seeds change separately.

\para{Legality.} A candidate that does not fit in 90\% of HBM or misses a latency target is rejected outright, and the ledger records why. Since the LLM's proposals go through the same checks and predictions as the seeds, it can waste trials but cannot return an illegal or unpriced configuration. For each (TP, PP, EP) shape, one anchor serves all scheduler and load trials. Where no anchor applies, the agent uses simulate mode, only for ranking.

\para{Kernel fusions.}\label{sec:fusion} An engine build can fuse adjacent kernels into one. The agent offers six fusion patterns found in kernel traces (Table~\ref{tab:fusion}). Each is priced at 75--100\% of its ideal saving, which is the launch cost of the removed kernels (Table~\ref{tab:constants}) plus the intermediate data no longer written to and read back from memory. Every trial reports the legal fusions saving at least 2\% of its decode step (Appendix~\ref{app:agent}).

\subsection{Mix planner}
\label{sec:mixplanner}

An operator serving several models needs a deployment mix, with one configuration per model and user class. The mix planner uses four latency tiers (Table~\ref{tab:mix}): realtime, interactive, responsive and bulk. Each tier sets a minimum per-user token rate (\emph{interactivity}) and a maximum TTFT. The planner predicts every candidate on the agentic trace with the DES. A candidate qualifies if it meets the interactivity target and its predicted TTFT is at most half the limit, which covers the TTFT error we measure in \S\ref{sec:variable}. Among those within 5\% of the best predicted throughput per GPU, it picks the one with most headroom.

\section{Evaluation}
\label{sec:eval}

We first check the parts that no anchor contains, communication and pipelines across nodes (\S\ref{sec:coll}--\S\ref{sec:ppnodes}). We then measure prediction error on fixed-length and agentic traffic (\S\ref{sec:fixed}--\S\ref{sec:variable}), check whether predictions order candidates correctly (\S\ref{sec:rank}), and test Pareto frontiers, speculative decoding and disaggregated serving (\S\ref{sec:pareto}--\S\ref{sec:disagg}). Finally we evaluate the planner's choices and its cost (\S\ref{sec:mix}).

\subsection{Methodology}
\label{sec:method}

\para{Setup.} All measurements use nodes of eight AMD Instinct MI355X GPUs (288\,GB of HBM3E each, connected by AMD Infinity Fabric\texttrademark{} links), with recent vLLM, SGLang and ATOM builds on ROCm\texttrademark{} 7 software and graph capture on. The models range from large MoE models (DeepSeek-V4-Pro and -Flash, GLM-5.2, Kimi-K3, MiniMax-M3, Qwen3.5) to smaller ones such as Qwen3-14B, gpt-oss-120b and Llama-3.1-8B~\cite{deepseekv4,glm5,kimik3,qwen3,gptoss,llama3}. Agentic runs replay the AgentX trace (\S\ref{sec:variable}), and 1k/1k means 1024 prompt and 1024 output tokens. Table~\ref{tab:sets} (Appendix~\ref{app:extra}) lists the sets.

\para{Metrics.} For fixed-length traffic we report the median absolute percentage error of mean TTFT, mean TPOT and output throughput. Agentic metrics vary by two orders of magnitude. For them we report an \emph{error factor}, the geometric mean of $\max(\hat y/y,\,y/\hat y)$ over runs, where $\hat y$ is predicted and $y$ measured. A factor of 1.00 is exact, and 1.10 means a typical 10\% miss in either direction. We report accuracy for anchored mode, and ranking for both modes.

\para{When an anchor covers a run.} An anchor \emph{covers} a target if it has the same engine build and the same layout (TP may be narrower), and was measured within $4\times$ of the target's concurrency. A narrower anchor must be a full batch sweep. Transferable choices such as prompt length, the KV tier, draft depth and replicas do not affect coverage. Covered targets use anchored mode. The rest are still projected in anchored mode from the nearest anchor of the same model, but only to rank, and are marked as such.

\subsection{Collective communication}
\label{sec:coll}

Communication outside the anchor's own GPUs, such as wider TP and EP groups and every transfer that leaves the node, comes from a model. We measured ROCm Communication Collectives Library (RCCL) collectives from 8\,KiB to 1\,GiB and vLLM's custom all-reduce from 4\,KiB to 8\,MiB, on 2--8 GPUs and across two nodes over AMD Pollara network cards. On points not used for fitting, every collective, group size and link type is within about 10\% mean error (Table~\ref{tab:coll}, Appendix~\ref{sec:comm-eval}).

\subsection{Pipelines and multiple nodes}
\label{sec:ppnodes}

A simple pipeline model predicts that TPOT does not change with the number of stages, but measured SGLang TPOT grows by 920--1130\,$\mu$s per stage. Each token passes every stage and returns from the last to the first, so \sys{} adds one transfer per stage, plus the host time that SGLang cannot hide under PP (Table~\ref{tab:constants}). On three models at PP 1--8 this gives 2.3--9.2\% median TPOT error, against 14.9--26.1\% for the simple model. We also predicted 24 two-node pipelines (one stage per node, TP1--8 in each, 2--16 GPUs) from single-node anchors without PP. Their median error is 11.6\% on TPOT and throughput (Appendix~\ref{app:ppmn}). The hop between nodes costs about the same as one inside a node, which makes PP a natural way to span nodes.

\subsection{Fixed-length workloads}
\label{sec:fixed}

\begin{figure}[t]
\centering
\includegraphics[width=\columnwidth]{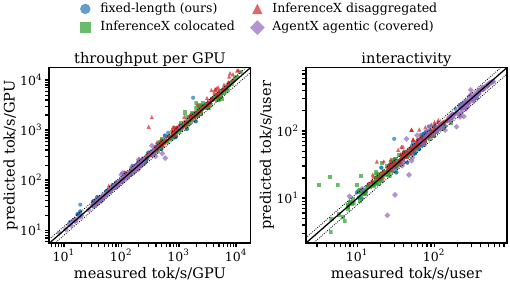}
\caption{Predicted against measured throughput per GPU and interactivity for every scored run that is not an anchor: our 172 fixed-length runs, 262 InferenceX runs (159 of them disaggregated) and the 95 covered AgentX runs, all in anchored mode. Dotted lines show $\pm$20\%.}
\label{fig:acc}
\end{figure}

\para{Setting.} We score 172 measured MI355X serving configurations of eight models with 1024- or 8192-token prompts, 1024 output tokens, 1--256 concurrent requests, TP 2--8 and three engines. Eleven anchors cover them all.

\para{Accuracy.} The median error is 8.1\% on TPOT, 8.8\% on throughput and 13.6\% on TTFT (Figure~\ref{fig:acc}, per model in Table~\ref{tab:fixed}, Appendix~\ref{app:extra}). Of all configurations, 93\% are within 20\% on TPOT and 88\% on throughput. TPOT and throughput are the most accurate because decode repeats one step that the anchor measures directly. Long prompts fare better: at 8192 tokens all three median errors are 6.3--6.9\%.

\para{Narrower anchors.} No anchor here uses more than half a node, so all 90 TP8 deployments are predicted from TP4 anchors. Their median throughput error is 9.7\%, against 6.1\% at the anchor's own width (Table~\ref{tab:tp4}, Appendix~\ref{app:narrow}).

\para{Tail latency.} On the 88 runs with at least 160 requests, the DES predicts the 99th percentile (p99) within a median 10.7\% on TPOT, 9.8\% on end-to-end latency and 13.8\% on ITL. For comparison, repeated runs of one configuration differ by 1--4\% on these p99s.

\subsection{Expert load under EP}
\label{sec:moe-eval}

\para{Routing is skewed, and prompts matter.} We replayed the recorded router of three MoE checkpoints at EP8. For DeepSeek-V4-Pro at batch 64 the busiest GPU receives 1.4--1.8$\times$ the mean routed tokens, against 1.2$\times$ for a uniform router. In a prefill chunk it receives up to 2.1$\times$, against 1.02$\times$. The prompt text matters as much as the batch. Random tokens, which fixed-length benchmarks send, give the hottest expert 46$\times$ its fair share in prefill, and real code spreads most evenly. Benchmarks on random tokens thus overstate production imbalance.

\para{Accuracy.} Nine vLLM runs shard experts (DeepSeek-V4-Flash at TP4/EP4, DeepSeek-V4-Pro at TP8/EP8), each predicted from an anchor without EP. With busiest-GPU factors from the recorded routing, TPOT is within 2.2\% and throughput within 2.1\% (1.4\% and 1.5\% with a sampled uniform router). On SGLang, without an all-to-all backend, expert outputs combine with the TP all-reduce. Pricing that all-reduce brings three held-out SGLang EP runs within 2.5\% with a uniform router and 10.5\% with the recorded routing (Appendix~\ref{sec:moe}).

\subsection{Variable-length agentic workloads}
\label{sec:variable}

\begin{table}[t]
\caption{Agentic AgentX replay on MI355X in anchored mode: error factor (1.00 is exact) on held-out configurations, predicted from 28 anchors. \emph{Covered} runs share an anchor's engine build, layout and precision and run within $4\times$ its session count. A run may appear in several rows. ITL plays the role of TPOT.}
\label{tab:variable}
\centering\scriptsize\setlength{\tabcolsep}{4pt}
\begin{tabular}{@{}lrrrr@{}}
\toprule
Held-out set & $n$ & Tput & ITL & TTFT\\
\midrule
Covered, all & 95 & 1.08 & 1.14 & 1.23\\
\quad $\le 2\times$ anchor's sessions & 38 & 1.07 & 1.11 & 1.19\\
\quad $2$--$4\times$ anchor's sessions & 57 & 1.08 & 1.16 & 1.26\\
\quad SGLang + HiCache & 12 & 1.15 & 1.40 & 1.38\\
\quad\quad host tier added to an anchor without one & 5 & 1.06 & 1.17 & 1.42\\
\quad ATOM + LMCache & 4 & 1.05 & 1.19 & 1.11\\
\quad vLLM + CPU offload & 4 & 1.04 & 1.04 & 1.25\\
\quad different draft depth & 2 & 1.08 & 1.06 & 1.15\\
\bottomrule
\end{tabular}
\end{table}

\para{Setting.} We replay SemiAnalysis's AgentX agentic coding trace~\cite{agentx} through the AIPerf client~\cite{aiperf}, with its recorded think times, on 184 MI355X configurations. They cover six models (DeepSeek-V4-Pro, DeepSeek-V4.1-Flash, GLM-5.2, Kimi-K3, MiniMax-M3, Qwen3.5), three engines, TP2--TP8 with and without EP and attention DP, 1--384 concurrent sessions, host KV tiers, several draft depths, routers and 13 disaggregated deployments.

\para{Anchors.} Each layout (model, engine build, precision and TP/EP/DP width) gets one anchor, fixed before any scoring. Among its runs without disaggregation or a host tier, it is the one closest to four sessions. No anchor is disaggregated. All runs use the model's published drafter. Every anchor therefore includes speculation, and only a different draft depth is transferred. Five layouts ran only with a host KV tier, so their anchors include it. The other 156 are held out.

\para{Accuracy.} On the 95 held-out runs that an anchor covers, anchored mode is within 8\% on throughput, 14\% on ITL and 23\% on TTFT (error factors 1.08, 1.14 and 1.23, Table~\ref{tab:variable}), across all six models and 1--384 sessions. Appendix~\ref{app:var} splits them by model, engine and precision.

\para{KV offload.} For the five SGLang runs whose host tier is added to an anchor measured without one, throughput is within 6\% and ITL within 17\%. Whether a turn's prefix is still in HBM, in DRAM or gone depends on an eviction order the anchor never sees. On two runs with a different draft depth, throughput is within 8\%.

\para{Decode context parallelism.} Thirteen runs use decode context parallelism (DCP), which splits each request's KV cache across GPUs during decode. They are GLM-5.2 on ATOM at TP4 with DCP4 and Kimi-K3 on vLLM at TP8 with DCP8. DCP changes the attention kernels and needs its own anchor, so each group is anchored on one of its own runs, not on a run without DCP. With anchors at the middle concurrency, the nine held-out runs are within 8\% (throughput) and 12\% (ITL).

We score the 13 agentic disaggregated runs in \S\ref{sec:disagg}.

\begin{figure}[t]
\centering
\includegraphics[width=\columnwidth]{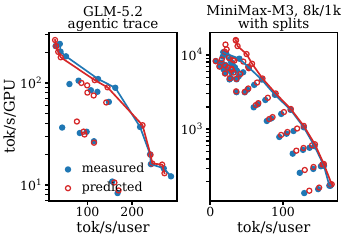}
\caption{Two measured Pareto charts. GLM-5.2 on the agentic trace keeps all ten measured-frontier configurations, and MiniMax-M3 at 8k/1k with disaggregated splits keeps 11 of 12. Measured runs are filled and anchored-mode predictions are open.}
\label{fig:frontiers}
\end{figure}

\subsection{Ranking fidelity}
\label{sec:rank}

For a search, getting the order right matters more than the absolute error. We group configurations that serve the same model, prompt length and concurrency into a \emph{choice set}, since these are the alternatives an operator compares. A pair is \emph{decisive} if its measured gap exceeds both the run-to-run noise and a floor of 3\%. We rank on throughput and TPOT (ITL on agentic traffic), not on TTFT. A predicted tie counts as half correct. Anchored mode orders 93.4\% of decisive throughput pairs and 85.4\% of TPOT pairs correctly (Table~\ref{tab:rank}, Appendix~\ref{app:extra}). Its throughput pick is the measured best in 73\% of sets and within 0.2\% of the best in nine of ten. Candidates in one set share an anchor, so their errors move together.

\para{Simulate mode} orders 89.0\% of throughput pairs and 84.3\% of TPOT pairs without any measurement. The fixed per-step cost is shared by candidates on one engine and mostly cancels, so simulate mode ranks within one engine.

\para{On agentic traffic.} We rank within one engine, because engines differ in per-step costs an anchor on another engine cannot see. The AgentX trace gives 41 such choice sets (same model, engine and session count) over 87 held-out runs. Anchored mode orders 90.9\% of decisive throughput pairs and all 51 ITL pairs, and its throughput pick is the measured best in 37 of 41 sets (simulate mode: 81.8\% and 90.2\%).

\subsection{Pareto frontiers}
\label{sec:pareto}

A deployment review usually asks for the trade-off between throughput per GPU and per-user interactivity, the Pareto frontier, for each model, engine and precision. \sys{} sweeps every engine, precision, parallel layout and concurrency, combined with scheduler settings, on 1k/1k and 8k/1k prompt/output lengths and on the AgentX trace, in both modes. Each drafter that InferenceX~\cite{inferencex} uses for a model gets its own line, at draft depths 1--8 with the best depth kept per concurrency. The sweep covers \ParModels{} models and \ParPoints{} points (\ParTracePoints{} on the trace) and runs on CPUs only.

\para{Measured frontiers.} We check frontiers only on measured configurations, which leaves \PMPoints{} configurations in \PMCharts{} charts, made up of our 172 fixed-length runs, 262 InferenceX fixed-length runs not used as anchors (\PMDisagg{} of them disaggregated), and the 95 covered AgentX runs. Anchored mode has \PMTputErr{} median error on throughput per GPU and \PMIntErr{} on interactivity, and a median Spearman correlation of \PMRho{} within a chart (Figure~\ref{fig:frontiers}, Table~\ref{tab:pareto}, Appendix~\ref{app:extra}). Of the \PMFront{} configurations on a measured frontier, \PMHit{} (\PMHitPct{}) are also on the predicted frontier. Deploying from the predicted frontier therefore loses no throughput per GPU at the median and \PMLossNinety{} at the 90th percentile, with at most 3\% lower interactivity.
\subsection{Speculative decoding}
\label{sec:spec}

The sweep has 43 measured speculative runs, all with MTP. EAGLE-3~\cite{eagle3}, DSpark and n-gram drafters are predicted only. Anchored mode predicts the MTP runs with 12.2\% median error on interactivity and 8.9\% on throughput per GPU, close to the 10.6\% and 9.8\% without speculation. At 50 tokens/s per user, the best speculative configuration gives a median $2.05\times$ the throughput per GPU of the best plain one (0.86--4.73$\times$ over ten charts). Its best draft depth ranges from 2 to 8 (median 5), so it must be searched per chart.

\subsection{Disaggregated serving}
\label{sec:disagg}

In a disaggregated deployment, prefill and decode run in separate pools of GPUs, and the KV cache is sent from one pool to the other. \sys{} predicts each pool with the step model for its phase, carrying a colocated anchor to the pool's layout. No disaggregated anchor is needed. Across several servers a cluster router (e.g., Dynamo~\cite{dynamo}, llm-d~\cite{llmd}) places requests; the DES routes each to the least-loaded prefill and decode replicas. Throughput per GPU counts the GPUs of both pools, and the KV transfer is priced on its backend (Appendix~\ref{app:disagg}). The sweep has eight pool layouts of 12--40 GPUs, including attention-DP pools (TP8, EP8, attention DP 8: ``DEP8''). Both kinds are swept to the same client counts, and five modeling details, among them each pool's own KV budget, decide which wins (Appendix~\ref{app:disagg}).

\para{Measured splits.} We evaluate on 159 disaggregated fixed-length deployments from InferenceX on MI355X. They cover DeepSeek-V4-Pro, GLM-5, MiniMax-M3 and Qwen3.5-397B at 1k/1k and 8k/1k, with TP4 or TP8 pools, 17 with attention DP. Pools use anchors from InferenceX's colocated runs. The median error is 5.5\% on throughput per GPU, 5.3\% on interactivity and 17\% on TTFT (Figure~\ref{fig:acc}, per backend in Table~\ref{tab:splits}, Appendix~\ref{app:disagg}).

\para{Where it pays, predicted.} In the GLM-5.2 sweep, disaggregated layouts give the most throughput per GPU between about 15 and 74 tokens/s per user, or up to about 170 with MTP. With 8k-token prompts, disaggregation often wins, by $1.14$--$2.72\times$, because long prefills no longer stall decode steps (Table~\ref{tab:disagg}). With 1k-token prompts colocation wins or ties every tier that both kinds meet. On the layouts InferenceX measured, the predicted winner is right in 19 of 21 tier decisions at 1k/1k and 23 of 27 at 8k/1k.

\begin{table}[t]
\caption{Colocated or disaggregated, by latency tier (anchored mode, no speculation, 14 charts). Predicted sweep: charts where the best disaggregated layout beats the best colocated one on throughput per GPU within the tier, and that gain at 8k/1k. Right call: on the layouts InferenceX measured, tier decisions where the predicted winner (split or colocated) is the measured one; no split is an anchor. $^*$Only a disaggregated layout meets the tier. The 1k/1k interactive win is $1.02\times$, a tie. The GLM-5.2 split in microscaling 4-bit floating point (MXFP4) with TP4 prefill and TP8 decode was predicted at $1.22\times$ in the interactive tier and measured in our own runs at $1.30\times$ with the split at the same 16 clients, or $1.67\times$ at 24, which the sweep skips.}
\label{tab:disagg}
\centering\scriptsize\setlength{\tabcolsep}{3pt}
\begin{tabular}{@{}lllc|cc@{}}
\toprule
 & \multicolumn{3}{c|}{Predicted sweep} & \multicolumn{2}{c}{Right call} \\
Tier (tok/s/user, TTFT) & 1k/1k & 8k/1k & 8k gain & 1k & 8k \\
\midrule
realtime ($\geq$150, $\leq$1\,s) & 1/1$^*$ & -- & -- & 2/2 & 2/2 \\
interactive ($\geq$70, $\leq$2.5\,s) & 1/4 & 3/4 & 1.37--2.72$\times$ & 4/5 & 6/7 \\
responsive ($\geq$25, $\leq$10\,s) & 0/7 & 7/7 & 1.19--1.58$\times$ & 6/7 & 7/9 \\
bulk (no limit) & 0/7 & 2/7 & 1.14--1.19$\times$ & 7/7 & 8/9 \\
\midrule
all & & & & 19/21 & 23/27 \\
\bottomrule
\end{tabular}
\end{table}

\para{On agentic traffic.} The AgentX data holds 13 measured splits (seven DeepSeek-V4-Pro on SGLang, six Kimi-K3 on vLLM) and no disaggregated anchor. Predicting each pool from a colocated anchor gives error factors of 1.12 on ITL and 1.08 on throughput (Appendix~\ref{app:disagg}). With these splits among the candidates, the planner picks a DeepSeek-V4-Pro DEP8 split in the responsive and bulk tiers. It is the measured best in one and 1.5\% short in the other; colocated-only picks lose 46\% and 43\%.

\subsection{Agent case studies}
\label{sec:mix}
\label{sec:search}

\para{Deployment mix: setting.} We gave the mix planner six models and, as candidates, the \MIXCONF{} AgentX runs an anchor covers (95 held out plus the 28 anchors), so every prediction is in anchored mode. Each candidate has a measured run. We can check each pick on hardware while the planner sees only predictions.

\para{Result.} In \MIXSLOTS{} of the 24 model--tier slots, at least one candidate meets the tier on hardware. Without using a GPU, the planner picks the measured best in \MIXEXACT{} of them and a configuration within \MIXBAND{} of the best in \MIXWTEN{} (Table~\ref{tab:mix}, Appendix~\ref{app:mixfull}), with a median throughput loss of \MIXREG{}. No exact pick is an anchor run.

\para{Short list.} A short list of the planner's five best predictions holds the measured best in 21 of the \MIXSLOTS{} slots and a configuration within \MIXBAND{} of it in 22, and every slot then has a candidate that meets the tier on hardware.

\para{Search scale.} The seed stage ran 192 searches (24 objectives, four models, with and without an SLO), scoring 56,328 candidates in 3--8 CPU-hours (Appendix~\ref{app:search192}).

\para{Cost.}\label{sec:cost} A prediction, including a DES run of $10\times C$ requests, takes 0.17--0.48\,s on one CPU core. The 226 fixed-length configurations anchored mode covers took 204 GPU-hours (671 runs) to measure, their 11 anchors 9.1 GPU-hours. The 172 scored in \S\ref{sec:fixed} are filtered separately. Simulate mode ranks 287 configurations (248 GPU-hours to measure) without a GPU.

\section{Related Work}
\label{sec:related}

\para{Profile-based serving simulators.} Vidur~\cite{vidur} profiles each operator for every model and TP width, and supports only dense models, so it cannot run any MoE model we evaluate. We ported it to MI355X and ran it on the 30 SGLang Llama-3.1-8B runs of \S\ref{sec:pp}. Its median errors are 34\%, 40\% and 40\% on TTFT, TPOT and throughput, against 14\%, 7.2\% and 7.4\% for \sys{}. Vidur times its own FlashInfer~\cite{flashinfer} kernels and models no host or per-stage overhead.

\para{Operator-level simulators with MoE.} APEX~\cite{apex} searches DP, PP, TP and EP plans from profiled operators. LLMServingSim~\cite{llmservingsim} co-simulates accelerators and serving software. AIConfigurator~\cite{aiconfigurator}, the closest to \sys{}, composes aggregated and disaggregated step times from a per-GPU kernel database. Frontier~\cite{frontier} and KernelSight-LM~\cite{kernelsight} add disaggregation, graphs, speculation or roofline kernel models to operator-level simulators. All build a step operator by operator and re-collect their databases per model, engine build and kernel library. \sys{} instead measures the whole step, so graph padding, fused kernels and host time come with the anchor (Table~\ref{tab:related}).

\para{Analytical and hardware-design models.} Roofline tools~\cite{llmviewer,genz}, Optimus~\cite{optimus}, LLMCompass~\cite{llmcompass}, Calculon~\cite{calculon} and ASTRA-sim~\cite{astrasim,astrasim2} compute step times from hardware descriptions, and BestServe~\cite{bestserve} ranks colocated and disaggregated strategies with per-stack efficiencies and no MoE. Like simulate mode they need no GPU, but they omit the fixed per-step cost that \S\ref{sec:transfer} keeps across widths.

\para{Tuning by trial.} SCOOT~\cite{scoot} tunes vLLM and TensorRT-LLM flags with Bayesian optimization over live runs, and LLM-guided optimizers~\cite{llambo,agenthpo} still pay for a real run per trial. \sys{}'s agent keeps the search loop~\cite{react,dspy} but replaces each trial with a sub-second prediction.

\para{Prediction from measurements.} Habitat~\cite{habitat}, Daydream~\cite{daydream}, Lumos~\cite{lumos} and hybrid measured-analytical models~\cite{primusneurips} predict training performance from measured kernels, traces or small benchmarks. Serving differs because the step also depends on traffic, through batching, prefix reuse, speculation~\cite{specdec} and disaggregation~\cite{splitwise,distserve}, which now also separates attention from experts~\cite{megascaleinfer} and prefill across datacenters~\cite{prfaas}.

\section{Limitations}
\label{sec:limits}

All measurements were taken on AMD Instinct MI355X GPUs. The method is not tied to this hardware: given short anchor runs of the same kernel sets on another GPU, the same transfer laws predict its other deployments. Extending the evaluation to NVIDIA GPUs is future work.

\section{Conclusion}
\label{sec:conclusion}

\sys{} predicts LLM serving performance from a few short measurements instead of a trial per configuration. It anchors each kernel set once, on at most one node, and carries the anchor to other deployments with transfer laws that keep the fixed per-step cost and scale only the work that more GPUs divide. Deployments larger than a node are reduced, measured and restored. A discrete-event simulator replays multi-turn agentic traffic, and a tuning agent and mix planner search with them. Eleven anchors measured in 9.1 GPU-hours predict fixed-length TPOT with 8.1\% median error and agentic throughput within about 8\%, and 159 additional disaggregated deployments within 5.5\% with no disaggregated anchor. Its predicted Pareto frontiers keep \PMHitPct{} of the measured frontier configurations, and its five-candidate short list reaches a near-best deployment in the feasible slots. Measuring each kernel set once and deriving the rest turns serving configuration from a search over hardware into a search over predictions.

\bibliographystyle{arxiv}
\bibliography{refs}

\section*{AI Usage Disclosure}
Generative artificial intelligence (AI) tools were used in this work as follows. Coding assistants in the Cursor editor, backed by large language models, were used to write and refactor parts of the predictor, the evaluation harness and the plotting scripts, to launch and monitor cluster jobs, and to analyze results; the authors reviewed the code, and every reported number comes from scripts run on recorded measurements (Appendix~\ref{app:repro}). The same assistants were used to search the literature and to draft and edit parts of the text, tables and figures of this paper; the authors revised all drafted text and verified the claims and citations against their sources. No AI tool generated measurement data. Separately, \sys{}'s tuning agent uses an LLM as part of the method (\S\ref{sec:agent}): the LLM only proposes configurations to try, and every performance number comes from the predictor, not from the LLM.

\amdattribution

\label{lastmainpage}
\clearpage
\appendix
\ifdefined\isarxivbuild\else
\setcounter{page}{1}
\onecolumn
\fi

\section{Design details}
\label{app:design}

\begin{table*}[h]
\caption{What \sys{} searches and how each choice is priced. $\dagger$: anchor-defining (changes the kernels, so it needs its own anchor); all other choices are transferred from an anchor or priced by the DES.}
\label{tab:space}
\centering\scriptsize\setlength{\tabcolsep}{3pt}
\begin{tabular}{@{}p{0.17\textwidth}p{0.40\textwidth}p{0.40\textwidth}@{}}
\toprule
Choice & Options & How it is priced \\
\midrule
\multicolumn{3}{@{}l}{\emph{Parallelism inside a replica}}\\
Tensor (TP) & 1--8 & sharded part $\propto 1/\mathrm{TP}$, unsharded floor kept (Eq.~\ref{eq:floor}) \\
Expert (EP) & divisors of experts & busiest rank gates the MoE layer; imbalance at both EPs \\
Attention data (DP) & divisors of TP & ranks own whole requests; DES runs parallel schedulers \\
Pipeline (PP) & divisors of layers, across nodes & $p$ hops plus host time PP stops overlapping (Eq.~\ref{eq:pp}) \\
Decode context (DCP) & 1--8 & KV and attention sharded; ranked only under a covering anchor \\
\midrule
\multicolumn{3}{@{}l}{\emph{Across replicas}}\\
Replicas & up to cluster size & DES routes sessions across replicas \\
Disaggregation & prefill and decode pools, each with TP, EP, DP, replicas; NVIDIA Inference Transfer Library (NIXL)~\cite{nixl}, Mooncake or MoRI & two DES pools; KV hand-off priced on the hand-off backend \\
\midrule
\multicolumn{3}{@{}l}{\emph{Kernels}}\\
Precision$^\dagger$ & weights, linear layers, experts, KV: bfloat16 (BF16), 8-bit floating point (FP8), 4-bit floating point (FP4) & anchor; KV dtype also sizes the pool \\
Backends$^\dagger$ & attention (aiter, Triton, Composable Kernel (CK)), sparse top-$k$, MoE, DeepEP~\cite{deepep}, fused kernels & anchor \\
Graph capture$^\dagger$ & none, piecewise, full & anchor; batch padded to captured sizes \\
Collectives & custom one-/two-shot, quick-reduce, fused root-mean-square normalization (RMSNorm), RCCL & $\alpha+m/B$ per variant (\S\ref{sec:comm}) \\
\midrule
\multicolumn{3}{@{}l}{\emph{Scheduling and load}}\\
Scheduler & running cap, token budget, chunk size, KV memory share & DES parameters (\S\ref{sec:knobs}) \\
Speculation & method$^\dagger$ (MTP, EAGLE-3, DSpark, n-gram), depth 0--8 & drafter cost and measured acceptance (\S\ref{sec:spec-sim}) \\
Memory tiers & host-DRAM KV tier, prefix cache & DES cache replay \\
Load & concurrency, arrival process and rate & DES \\
\bottomrule
\end{tabular}
\end{table*}

Parallelism forms are mostly transferable, while kernel choices are anchor-defining. A deployment can combine TP, EP, attention DP, PP inside or across nodes, decode context parallelism, replicas behind a router, and disaggregated prefill and decode pools joined by a KV hand-off. Under attention DP each rank owns whole requests, so a TP8 replica does not store a latent cache eight times. These forms change which GPU does which work, not which kernels run, so they are transferable. The exception is DCP, which changes the attention kernels and is ranked only under a covering anchor (\S\ref{sec:variable}). Precisions, backends, graph capture and the speculative method change the kernels, so each needs its own anchor.

\subsection{Discrete-event simulator}
\label{app:des}

\para{Input.} The DES takes either a synthetic workload or a trace. A synthetic workload gives prompt and output lengths (fixed or drawn from a range), an arrival process, and optionally a pool of shared prefixes whose popularity is uniform or Zipf-skewed. A trace follows the Mooncake format, as JSON lines or a JSON array. Each record has an arrival timestamp, input and output lengths, and \texttt{hash\_ids}, the ordered hashes of the prompt's fixed-size token blocks. Optional fields add a think time, a session id, the request this turn waits on, and the requests it joins (a turn that waits for several sub-agent calls). A compact form lets a record reuse the first $n$ blocks of an earlier record, so long agentic contexts are not repeated in the file. Hash blocks are 512 tokens in Mooncake's traces and 64 in AgentX; the DES infers the size from the trace and counts cache capacity in these blocks.

\para{Prefix cache.} Each replica owns a content-addressed block cache. A block's hash covers its whole prefix, so a block also identifies its predecessor and the cache forms a tree, as in a radix cache. A request's hit is the longest leading run of its blocks already resident; only the uncached suffix is prefilled. Its blocks are then inserted, and blocks beyond the capacity the memory model gives the KV pool are evicted. Eviction takes the least recently used (LRU) \emph{leaf}: a block is never dropped while a resident block continues from it, so shared heads survive pressure and only divergent tails are reclaimed. This matters: on the AgentX trace at 256 clients against DeepSeek-V4's 231,336-block pool, flat LRU turns the sweep over contexts into a cycle that evicts every block just before its reuse (0\% hits), while leaf-first eviction gives 90.6\%, against the trace's published reuse of about 92\%. With a host-memory tier, evicted blocks move one tier down instead of being dropped, and a hit found there makes the request wait for the load.

\para{Admission order.} Hits depend on the order requests reach the cache, so the cache is replayed in the order the engine admits them. Requests that fit in the KV pool run in arrival order. When a backlog builds up, the engine picks the waiting request with the longest resident prefix (SGLang's default policy, which vLLM's prefix-caching scheduler approximates), and the replay does the same over the backlog only. A session turn is not a candidate until the request it waits on has been served, so it cannot take the hit that request creates. Each client's next turn refills that client's slot, so the context it continues is the one just served.

\para{Routing.} With several replicas, the router picks a replica for each request before its hit is resolved. Policies are least-loaded, a KV-aware cost that trades prefix overlap against load like a cluster router's, prefix hashing that colocates requests sharing a leading block, round robin and random. Under disaggregation a request goes to a prefill replica (the one holding its prefix on traces), its KV is handed off on the transfer backend, and it decodes on the least-loaded decode replica.

\para{Timed simulation.} Each replica then runs its own event loop over a virtual clock: arrivals, scheduler steps and completions. Closed-loop clients send their next request when the previous one finishes plus its think time; open-loop requests arrive at their timestamps. At each step the scheduler model fills the batch as the target engine does, within the running-request cap and token budget, with chunked prefill, mixed or prefill-only steps, and pages of KV per request. A request's cached prefix counts as already computed. The step cost model (anchored or simulate mode) prices the step's mix of prefill tokens and decode contexts, the clock advances by that time, and each running request emits a token. Host time per request is added to TTFT and end-to-end latency (\S\ref{sec:des}). Attention-DP ranks run as parallel schedulers in lockstep, so a step costs as much as its slowest rank.

\para{Output.} Per-request TTFT, ITL, TPOT and end-to-end latency give medians and tails, and their totals give throughput per GPU and the hit rate per replica. The tuning agent receives these numbers for each candidate. They come from the same step cost model the analytical predictor uses, so the two agree when queueing and cache effects are absent, and the DES adds those effects when they are present.

\subsection{Speculative decoding model}
\label{app:spec}

An anchor measured with speculation already contains the draft cost and the acceptance. Elsewhere \sys{} models the drafter. MTP and EAGLE-3 heads are built from the target's own layer types, so each drafter layer is priced as a target layer of the same kind, plus the language-model (LM) head and sampling, at the drafter's vocabulary and context. The number of drafter layers is a parameter: MTP defaults to the model's MTP layer count and EAGLE-3 to one dense layer over a pruned vocabulary, and a model card or flag can change either. DSpark~\cite{dspark} defaults to a three-layer MoE backbone per block plus a light serial head, and n-gram drafting is free. In anchored mode the draft cost is the measured step scaled by the modeled ratio of draft time to step time. A verify step at depth $k$ commits
\begin{equation}
L(k) = 1 + \sum_{i=1}^{k} \prod_{j=1}^{i} a_j
\label{eq:al}
\end{equation}
tokens, where $a_j$ is the probability that draft token $j$ is accepted given that tokens $1..j{-}1$ were. For models where InferenceX~\cite{inferencex} publishes measured acceptance lengths, \sys{} takes $L(k)$ from them. Other models use the sibling head's curve (MTP for EAGLE-3 and vice versa) or the drafter's default, the per-depth median over InferenceX models. Beyond the last measured depth, $L$ is extrapolated with a geometric acceptance chain fitted to the measured points. Acceptance also depends on how predictable the text is. SPEED-Bench~\cite{speedbench} measures it in six domains, which \sys{} groups into three optional entropy classes: low (coding, math), mixed (science, question answering) and high (writing, roleplay). The measured curves count as coding, and a class scales the first-position acceptance by its mean rate relative to coding. When the depth is left open, \sys{} tries every depth from 0, meaning no speculation, up to the limit and keeps the one with the most tokens per unit of step time at the projected batch, so it speculates only when that pays. On 43 measured MTP runs, anchored mode with the drafter model is nearly as accurate as on runs without speculation (\S\ref{sec:spec}).

\section{Component models}
\label{app:models}

\begin{table}[h]
\caption{Measured constants on MI355X. Collectives cost $\alpha+m/B$; ranges span 2--8 GPUs. Each is measured once per platform and reused for every model.}
\label{tab:constants}
\centering\scriptsize\setlength{\tabcolsep}{3pt}
\begin{tabular}{@{}lll@{}}
\toprule
Component & Constant & Value\\
\midrule
Custom all-reduce, in node & $\alpha$, $B$ & 4.9--11.0\,$\mu$s, 56--123\,GB/s\\
RCCL all-to-all, in node & $\alpha$, $B$ & 35--38\,$\mu$s, 113--409\,GB/s\\
Send/recv, two nodes & $\alpha$, $B$ & 44\,$\mu$s, 45\,GB/s\\
All-reduce, 16 GPUs, two nodes & $\alpha$, $B$ & 91\,$\mu$s, 32\,GB/s\\
Kernel occupancy (fusions) & per kernel & 4.98\,$\mu$s\\
Pipeline hop (SGLang) & per stage & 970\,$\mu$s\\
Host time PP stops overlapping & per step & $0.33+0.0063\,b$\,ms\\
Tokenization & per prompt token & 5--11\,$\mu$s\\
KV hand-off, MoRI & floor, $B$ & 120\,$\mu$s, 47.9\,GB/s\\
\bottomrule
\end{tabular}
\end{table}

\subsection{Decode shard per doubling}
\label{app:shard}

Doubling TP does not always halve the excess above the floor in Eq.~\ref{eq:floor}, so \sys{} measures the shrink factor $\rho$ when it can. This needs the same stack to have been harvested also at half the anchor's width. $\rho$ is then the ratio of the excess at $w$ to the excess at $w/2$, at each batch both sweeps share. It is clamped to $[0.5,1]$, interpolated in $\log b$, and applied once per doubling from $w$ to $w'$. Both harvests stay at or below the anchor's width, so the target width is still never measured. Only the decode half of the sibling harvest is used, because the prefill transfer already has its own length curve. Per doubling, the excess shrinks to 0.50, the ideal, for DeepSeek-V4-Pro on ATOM (TP4 to TP8). It shrinks only to 0.55--0.69 for GLM-5.2 on SGLang (TP2 to TP4 and TP4 to TP8), 0.68--0.71 for DeepSeek-V4-Flash on SGLang and 0.76 for MiniMax-M2.7 on ATOM (both TP2 to TP4).

\subsection{Communication}
\label{sec:comm}

Per layer, the kernel charges two TP all-reduces (one under attention DP). Per MoE layer, it charges an EP dispatch and combine. Per step, it charges $p$ point-to-point hops across pipeline stages (\S\ref{sec:pp-model}). Every collective costs a latency plus a bandwidth term, $t=\alpha+m/B$, on one of two tiers.

\para{Inside a node.} Decode messages are small. Below 8\,MiB, vLLM does not call RCCL~\cite{rccl}; it uses its own one- and two-shot custom all-reduce. We price that range with $\alpha+m/B$ measured on the custom kernel, per group size and per kernel variant. All-to-all below 32\,MiB is priced the same way on RCCL. Inside a step we charge only $m/B$, because $\alpha$ is the per-kernel launch and occupancy cost that the step already charges for every kernel it issues. We measured that cost as 4.98\,$\mu$s per kernel from a TP1--8 sweep. Large messages go to RCCL over the xGMI mesh. In a fully connected node of $N$ GPUs, each GPU has $N{-}1$ links and a group of $g$ GPUs uses $g{-}1$ of them, so $B(g)=B_\text{node}\,\eta\,(g{-}1)/(N{-}1)$ with one node parameter $B_\text{node}$ and an efficiency $\eta$. On MI355X, $N=8$. In anchored mode, the anchor step already contains the engine's communication, and transfer moves it with the step.

\para{Across nodes.} Traffic that leaves the node crosses the network interface cards (NICs). We price it from the NIC datasheet bandwidth and a per-message latency, and hierarchical collectives combine the two tiers. Pipeline stages are placed contiguously, so a pipeline over $n$ nodes has $n{-}1$ forward boundaries on the network, plus the token return from the last stage to the first. All other hops stay on xGMI. \S\ref{sec:comm-eval} validates both tiers on one and two nodes.

\subsection{MoE expert load}
\label{sec:moe}

Under EP, an MoE layer ends only when its most loaded rank ends, because each rank runs only its own experts. Two effects make ranks uneven. The first is chance. A decode step at batch $b$ makes only $bk$ top-$k$ choices, so how many land on each rank varies from step to step. The second is skew. A trained router favors some experts whatever the input, and the tokens of one prompt pick alike. Chance is large in inference decode, which runs at small batch. Training steps carry thousands of tokens, so there it averages out.

\para{The busiest rank gates each layer.} For a step of $t$ tokens at width EP, \sys{} uses two factors: $I_n=\mathbb{E}[\max_r n_r]/\bar n$, the busiest rank's routed tokens over the mean, and $I_d=\mathbb{E}[\max_r d_r]/\bar d$, its distinct experts read over the mean. Both are expectations of a per-layer maximum, because each layer waits for its own busiest rank. The busiest rank's expert GEMM is priced with $I_n\bar n$ rows over $I_d\bar d$ experts: at weight-bound decode the experts read set the cost, and at compute-bound prefill the rows do. Token permutation and the activation scale with $I_n$; attention, the router and communication are unchanged. At $\mathrm{EP}=1$ every rank holds a slice of every expert, so both factors are 1. With no measurement, \sys{} samples a uniform top-$k$ router, which captures chance and not skew. A uniform router at EP8 already gives $I_d=1.75$ at $b=4$, $1.31$ at $b=16$ and $1.09$ at $b=64$ for GLM-5.2's 256 experts and top-8, and under $1.08$ on rows once a step carries 512 tokens or more.

\para{Skew is measured from the checkpoint's routing.} We serve the checkpoint on SGLang at TP8 with its expert-distribution recorder on, which logs the top-$k$ choice of every token in every layer, and replay the log on any EP layout offline. A decode batch of $b$ requests is $b$ tokens of one recorded decode step, each from a different request; a prefill chunk is a contiguous run of prompt tokens. One recording takes a few minutes of serving once the weights are loaded and covers every EP width and batch; the replay is stored as a table of $I_n$ and $I_d$ per EP, phase and token count, which both modes read with no GPU: simulate mode prices the busiest rank's expert GEMM with it directly, and anchored mode takes its ratio between the anchor's and the target's EP. Engines can mitigate imbalance by replicating hot experts (EPLB); the replay prices that too, by placing $R$ extra copies of the hottest experts greedily on long-run load. Without a recording, \sys{} keeps the analytical bound in which only the share of routed mass that still lands on a single slot stays hot:
\begin{equation}
I' = 1 + (I-1)\,\frac{E}{E+R}.
\label{eq:eplb}
\end{equation}

\begin{table}[t]
\centering
\small
\caption{Busiest-rank routed tokens over the mean ($I_n$) for DeepSeek-V4-Pro (384 experts, top-6) at EP8, replayed from recorded routing. Random tokens are what fixed-length benchmarks send, sonnet text is what the AgentX replay sends, and code is real Python source. Uniform is a sampled uniform router (chance only); EPLB adds 48 replicas.}
\label{tab:moe-load}
\begin{tabular}{lcccc}
\toprule
Prompts & decode $b{=}4$ & decode $b{=}64$ & decode, EPLB $b{=}64$ & prefill 8k \\
\midrule
Uniform router & 1.89 & 1.20 & -- & 1.02 \\
Random tokens  & 2.03 & 1.60 & 1.41 & 2.06 \\
Sonnet text    & 2.07 & 1.77 & 1.17 & 1.76 \\
Code           & 1.95 & 1.41 & 1.19 & 1.44 \\
\bottomrule
\end{tabular}
\end{table}

\para{What the recording shows.} Table~\ref{tab:moe-load} gives DeepSeek-V4-Pro at EP8. At $b\le 4$ the real router is within 10\% of the uniform one: chance dominates. From $b\approx 16$ skew dominates, and it does not average out: the busiest rank carries $1.4$--$1.8\times$ the mean at $b=64$, and up to $2.1\times$ in a prefill chunk, against $1.2\times$ and $1.02\times$ for a uniform router. The prompts matter as much as the batch. Random tokens, which fixed-length benchmarks send, concentrate prefill on few experts, and the hottest expert takes $46\times$ its fair share of routings; real code spreads the most evenly. A projection should therefore use the recording whose prompts resemble the target's traffic, and a benchmark scored on random tokens overstates the imbalance of production traffic. At $b=64$ replication removes most of the skew on sonnet text and code but less than half on random tokens, and none of the chance at small batch, because it balances long-run load, not a single step. The other two recorded checkpoints follow the same pattern at EP8: at $b=64$ the busiest rank carries $1.30$--$1.66\times$ the mean on GLM-5.2 and $1.44$--$1.89\times$ on DeepSeek-V4-Flash, lowest on code in both, and in an 8192-token prefill $1.30$--$1.74\times$ and $1.47$--$2.14\times$.

\para{Imbalance is evaluated at both widths.} An anchor is harvested at reduced width (\S\ref{sec:anchors}), often with a smaller EP than the target. If one imbalance value were applied to both sides of the transfer ratio, it would cancel, and a wider EP would look free. \sys{} therefore evaluates $I_n$ and $I_d$ separately at the anchor's EP and at the target's EP, so the ratio keeps the penalty of sharding experts more finely. Training-side projection reduces EP for the same reason and restores it analytically. On the nine measured runs in our corpus that shard experts (DeepSeek-V4-Flash on vLLM at TP4/EP4, 2--256 clients, and DeepSeek-V4-Pro at TP8/EP8), each restored from an anchor harvested without EP, anchored mode projects TPOT within a median 2.2\% and throughput within 2.1\% with the recorded random-token routing, which is what these benchmarks send, and within 1.4\% and 1.5\% with the uniform router.

\para{A controlled pair.} The corpus runs change EP together with the anchor's width. To isolate EP, we served DeepSeek-V4-Pro on SGLang at TP8 on one node twice, once with every rank holding a slice of every expert (EP1) and once with experts sharded (EP8), at 1, 4, 16 and 64 clients on 1k prompts and 4 and 16 on 8k. EP8 is 10--13\% slower in TPOT up to 16 clients and level at 64. Projected from the EP1 run, anchored mode reads EP8's TPOT within a median 3.5\% and its throughput within 3.4\% without the busiest-rank factors. On this engine the step pays for reading fewer, wider experts, while on vLLM it also pays for the busiest rank.

\para{SGLang's EP dispatch.} With \texttt{-{}-ep-size} and no all-to-all backend, SGLang keeps tokens replicated across the TP group and combines expert outputs with the TP all-reduce. Pricing an EP all-to-all on top of the all-reduce charges about 19\,ms per prefill forward. With dispatch priced as SGLang runs it, the pair above fits best with half of the busiest rank's surplus of distinct experts ($I_d$) kept, and TPOT is within 3.0\%. We then scored three SGLang runs that change only \texttt{-{}-ep-size} against the same configuration without EP, none of them used in the fit: GLM-5.2 MXFP4 at EP4 and EP8 on 1k prompts and DeepSeek-V4-Pro at EP8 on 8k prompts, all at 64 clients. Priced with the all-reduce and both factors in full they are within $-2.2$ to $+2.5\%$ under the uniform router and at most $+10.5\%$ under the recorded routing.

\para{Few experts read slowly.} A grouped GEMM over few experts reads HBM below peak, and \sys{} charges the shortfall. One or two experts give too few independent streams to saturate HBM, and how far they fall short depends on the kernel. On MI355X at one row per expert, one expert reaches 25\% of peak bandwidth with DeepSeek's FP8 kernel and 14\% with gpt-oss's MXFP4 kernel. Both saturate by about 16 experts. \sys{} charges the shortfall for the group size from the curve of the expert dtype. Sharding by TP cuts the bytes without cutting the group, so it is not charged. This term decides EP transfer at decode, where an EP8 rank reads one to three whole experts instead of slices of all of them.

\para{The anchor's own router.} In anchored mode the router's skew at the anchor's width is inside the measured step, because the harvest loads real weights; random weights spread tokens evenly and make an MoE anchor too fast (\S\ref{sec:lessons}). What the anchor cannot hold is the imbalance at a wider EP, which is why the factors above are evaluated at both widths. Simulate mode also needs how many distinct experts a decode step reads, because a small-batch MoE step is bound by weight bandwidth and its cost follows the experts touched, not the experts held. For $t$ tokens with top-$k$ routing, independent routing touches a fraction $1-(1-k/E)^t$ of the experts. Real tokens touch fewer. \sys{} takes this from a coverage curve measured on the checkpoint's own router: for DeepSeek-V3 (256 experts, top-8) it bottoms at 0.89 of the independent prediction at 64 tokens, while for gpt-oss-120b (128 experts, top-4) it bottoms at 0.78. The gpt-oss curve is fitted to decode latency, so it also absorbs the grouped-GEMM efficiency the kernel model misses, and is an effective coverage rather than a physical one.

\subsection{Pipeline parallelism}
\label{sec:pp-model}

Serving has no backward pass, so a PP layout runs every layer of a token in sequence, and the only overlap comes from micro-batches in flight. A serial model (the sum of the stages plus $p{-}1$ activation transfers of a few KiB) predicts that TPOT does not depend on depth. Measured TPOT on SGLang grows steadily with depth (\S\ref{sec:pp}). Three measured facts explain why.

\para{Every step makes $p$ hops.} At one client, the added TPOT per stage is 920--1130\,$\mu$s on three models at PP 2, 4 and 8 (Figure~\ref{fig:pp}a), about 20$\times$ an eager send/recv of the activation. A step makes $p{-}1$ activation sends plus one more hop: the last stage returns the sampled token to stage 0 before the next step can start. \sys{} charges $p\,(t_\text{p2p}+o)$ per step, with one per-hop overhead $o=970\,\mu$s.

\para{Scheduler host time stops overlapping.} On one stage, SGLang overlaps CPU scheduling of step $k{+}1$ with the GPU forward of step $k$. With PP$>$1, it turns that overlap off. Running the same single-stage engine with overlap on and off isolates the host time $H(b)$. On all three models it is 0.31--0.39\,ms at small batch and 0.69--0.78\,ms at 64 requests, and $H(b)=0.33\,\text{ms}+6.3\,\mu\text{s}\cdot b$ fits well (Figure~\ref{fig:pp}b).

\para{In a full pipeline, host time adds up.} The clients split into $p$ micro-batches of $C/p$. With two or more micro-batches, no stage waits for another, so each of the $p$ stages adds its host time to the cycle:
\begin{equation}
s_\text{PP}(C) = s\bigl(C/p\bigr) + p\,(t_\text{p2p}+o) + p\,H\bigl(C/p\bigr),
\label{eq:pp}
\end{equation}
where $s$ is the full-depth forward at the micro-batch size. Nothing here is fitted to a PP ladder: $o$ comes from concurrency-1 runs and $H$ from single-stage probes.

\para{Across nodes.} For hops that cross the network, $t_\text{p2p}$ is the NIC-tier send/recv of \S\ref{sec:comm}. The per-hop overhead $o$ and host time $H$ are engine costs and do not change. So a multi-node pipeline projects from the same single-node anchors (\S\ref{sec:multinode}).

\subsection{Memory and capacity}
\label{sec:memory}

The memory model is fully analytical. It covers weights by component dtype and layout; KV bytes per token, including sparse-attention indexer keys and compressed pools; and the activation working set. It sets the largest sustainable concurrency, so \sys{} never reports throughput at a concurrency that HBM cannot hold. Against the pool size each engine prints at startup, projected KV bytes per token agree to 0.0\% for GLM-5.2 (BF16, 49 logs) and MiniMax-M3 (FP8), and to 2\% for GLM-5.2 FP8. Some pools do not reduce to per-token arithmetic, such as hybrid linear attention and DeepSeek-V4's compressed multi-head latent attention (MLA) pool. For those we use the size the engine reports, so admission is always simulated against the real pool.

\subsection{KV cache below HBM}
\label{sec:kv-tier}

The DES models the tier as a second level of its block cache. Blocks evicted from HBM by the engine's leaf-first LRU rule move to DRAM and are dropped only when DRAM is also full. A request whose prefix is DRAM-resident has those blocks promoted back, and the DES queues the transfer on the host link of its rank before scheduling the request, so a load adds to that request's TTFT and does not stall other requests' steps; loads that arrive together queue behind each other. The tier's size is the DRAM each run hands its engine, divided over its GPUs (256--2{,}399\,GB per node here), and the link is rated at the host-to-device bandwidth. Hits are therefore an output of the replay: a tier that is too small for the sessions in flight recomputes them, as the engine does. The four offload engines share this model and differ in their DRAM budget and in the scheduler they sit under. Covered offloaded runs are predicted within 4--16\% on throughput per engine, and within 6\% where the tier is added to an anchor measured without one (\S\ref{sec:variable}).

In the Pareto sweep the tier is an input, its size per GPU and its link bandwidth, and the frontier is drawn with KV in HBM only and with a DRAM tier.

\section{Pipeline parallelism and multiple nodes}
\label{app:ppmn}

\subsection{Pipeline parallelism}
\label{sec:pp}

\begin{table}[t]
\caption{Pipeline parallelism on SGLang: median absolute TPOT error (\%) over the PP$>$1 rows, for the serial model and the per-stage model of Eq.~\ref{eq:pp} (this work).}
\label{tab:pp}
\centering\scriptsize
\begin{tabular}{@{}l r rr@{}}
\toprule
Model & rows & serial & \textbf{this work}\\
\midrule
Llama-3.1-8B & 24 & 18.1 & \textbf{7.2}\\
Qwen3-14B-FP8 & 9 & 14.9 & \textbf{2.3}\\
Qwen3.6-35B-A3B & 9 & 26.1 & \textbf{9.2}\\
\bottomrule
\end{tabular}

\end{table}

\begin{figure*}[t]
\centering
\includegraphics[width=\textwidth]{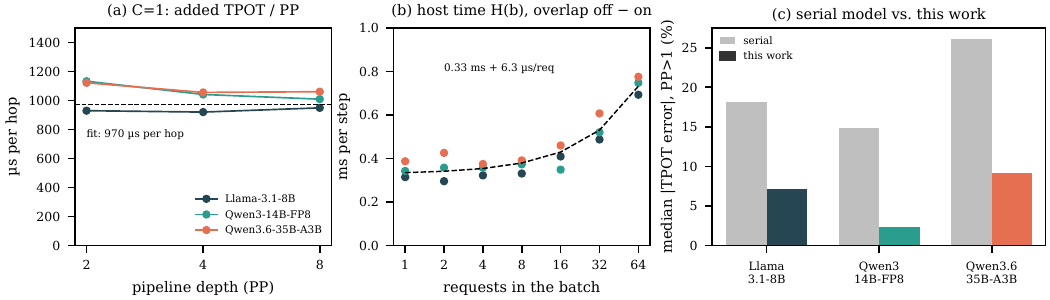}
\caption{Pipeline parallelism on SGLang, three models. (a) At one client, the added TPOT per stage does not change with depth: 920--1130\,$\mu$s, because every step makes $p$ hops. (b) Host time per step (overlap-off minus overlap-on, one stage), against $0.33\,\text{ms}+6.3\,\mu\text{s}\cdot b$. (c) Median TPOT error over PP$>$1 rows for the serial model and Eq.~\ref{eq:pp}, anchored mode.}
\label{fig:pp}
\end{figure*}

We measured three models on SGLang: Llama-3.1-8B-Instruct, Qwen3-14B-FP8 and Qwen3.6-35B-A3B, at TP1 with PP 1, 2, 4 and 8, concurrency 1, 16 and 64, 1024-token prompts and outputs, and prefix caching off. Llama also has 8192-token prompts and a TP2$\times$PP2 layout. Each model's PP layouts are projected from its PP1 anchor, so the study isolates the PP model. Appendix~\ref{app:pp} gives absolute values and signed errors.

\para{Accuracy.} Eq.~\ref{eq:pp} has 7.2\%, 2.3\% and 9.2\% median TPOT error on Llama, Qwen3-14B-FP8 and Qwen3.6-35B-A3B; the serial model has 18.1\%, 14.9\% and 26.1\% (Table~\ref{tab:pp}, Figure~\ref{fig:pp}c). Nothing in Eq.~\ref{eq:pp} was fitted to these ladders.

\para{Inside a node, PP buys capacity, not speed.} At one client and 1024-token prompts, Llama TPOT rises from 4.46\,ms at PP1 to 6.32, 8.15 and 12.06\,ms at PP2, 4 and 8 (4.57 to 7.18, 9.18 and 14.24\,ms at 8192 tokens). At 1024 tokens and 64 clients, throughput falls from 7,773 to 6,988, 5,809 and 3,833\,tok/s. At 8192 tokens and 64 clients, PP raises absolute throughput from 2,519 to 3,151--3,494\,tok/s, but uses 2--8$\times$ the GPUs: PP4 gives 874\,tok/s per GPU against 2,519 for PP1. PP is for KV capacity and for models too large for one layout. Its most important use is spanning nodes (\S\ref{sec:multinode}).

\subsection{Serving across nodes}
\label{sec:multinode}

\begin{figure}[h]
\centering
\includegraphics[width=\columnwidth]{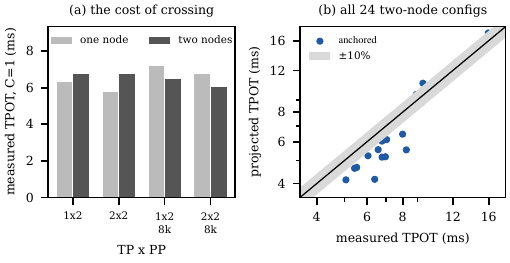}
\caption{SGLang Llama-3.1-8B pipelines across two MI355X nodes. (a) Measured TPOT at one client for layouts run both on one node and across two: crossing the NICs adds no systematic cost. (b) Projected against measured TPOT for all 24 two-node configurations, from single-node anchors; the band is $\pm$10\%.}
\label{fig:mn}
\end{figure}

\begin{table}[t]
\caption{SGLang Llama-3.1-8B pipelines split across two MI355X nodes: median absolute error (\%) per layout, projected from single-node anchors.}
\label{tab:mn}
\centering\scriptsize
\begin{tabular}{@{}lrrrr@{}}
\toprule
 & GPUs & & & \\
Layout & /node & rows & TPOT & Tput\\
\midrule
TP1$\times$PP2 & 1 & 6 & 6.7 & 6.3\\
TP2$\times$PP2 & 2 & 6 & 9.5 & 11.3\\
TP4$\times$PP2 & 4 & 6 & 13.7 & 14.9\\
TP8$\times$PP2 & 8 & 6 & 21.7 & 33.8\\
\midrule
All & & 24 & 11.6 & 11.6\\
\bottomrule
\end{tabular}

\end{table}

We split the SGLang Llama-3.1-8B pipelines of \S\ref{sec:pp} across two MI355X nodes in the standard layout: one pipeline stage per node, TP inside each node, and the stage boundary and token return on the Pollara NICs. Four layouts, TP1 to TP8 per node (2 to 16 GPUs), run at 1024- and 8192-token prompts and 1, 16 and 64 clients: 24 configurations. The projection uses the same single-node PP1 anchors as \S\ref{sec:pp} and no two-node measurement. The placement-aware pipeline model moves the two cross-node hops onto the NIC tier; nothing else changes.

\para{Results.} Anchored mode projects the two-node pipelines at 11.6\% median error on both TPOT and throughput (Table~\ref{tab:mn}, Figure~\ref{fig:mn}b), for layouts of up to 16 GPUs, including TP4 and TP8 stages never measured under a pipeline.

\para{Crossing nodes is cheap.} The model and the measurements agree that crossing nodes adds little. Inside a step the model prices each NIC hop at a few microseconds of wire latency plus one kernel launch, not the 44\,$\mu$s $\alpha$ of a standalone send/recv benchmark (Table~\ref{tab:constants}), for the same reason as inside a node (\S\ref{sec:comm}). Over the four layout and prompt-length pairs that also ran on one node, that is at most 10\,$\mu$s per token, against a 970\,$\mu$s per-hop engine cost. At one client, the same layout on one node and on two differs by $-$0.69 to $+$1.03\,ms per token, with no consistent sign, and two of the four pairs ran faster across nodes (Figure~\ref{fig:mn}a). So pipeline parallelism is the natural way to span nodes.

\para{What it means for search.} Every two-node candidate here was projected from single-node anchors in under half a second each, while measuring the two-node ladder held 16 GPUs for about an hour and a half. The larger the deployment, the larger the share of the space that needs no measurement.

\section{Additional results}
\label{app:extra}

\begin{table*}[h]
\caption{Evaluation sets, all measured on MI355X. Engines: A=ATOM, S=SGLang, V=vLLM. Configurations are measured ones, except for the Pareto sweep, which is projected and compared with every measured run it matches. Anchors are the distinct anchors each set is projected from. Model versions differ between sets (DeepSeek-V4-Flash and MiniMax-M2.7 in the fixed-length corpus, DeepSeek-V4.1-Flash and MiniMax-M3 in the AgentX replay, GLM-5 in InferenceX and GLM-5.2 in our own runs); each name is the checkpoint that set measured.}
\label{tab:sets}
\centering\scriptsize\setlength{\tabcolsep}{4pt}
\begin{tabular}{@{}p{2.3cm}clp{5.6cm}rrp{3.4cm}@{}}
\toprule
Set & Models & Engines & Traffic & Configs & Anchors & Scored on \\
\midrule
Pipelines & 3 & S & 1k or 8k in, 1--64 clients, PP 1--8, one or two nodes & 24 (2 nodes) & 4 + 4 & pipeline model (\S\ref{sec:ppnodes}) \\
Fixed-length corpus & 8 & A, S, V & random prompts, 1k or 8k in, 1k out, 1--256 clients & 172 & 11 & error and order (\S\ref{sec:fixed}, \S\ref{sec:rank}) \\
Agentic replay & 6 & A, S, V & AgentX coding trace, 1--384 sessions, with and without KV offload, colocated and split & 184 & 28 & error where covered, order, deployment mix (\S\ref{sec:variable}, \S\ref{sec:mix}) \\
Disaggregated splits & 4 & A, S, V & 1k or 8k in, 1k out, 1--2{,}048 clients, 8--24 GPUs & 159 & 16 & error with no disaggregated anchor (\S\ref{sec:disagg}) \\
Pareto sweep & 13 & A, S, V & 1k/1k, 8k/1k and AgentX & 169{,}962 proj. & 23 & frontier error and order (\S\ref{sec:pareto}) \\
\bottomrule
\end{tabular}
\end{table*}

\begin{figure*}[h]
\centering
\includegraphics[width=\textwidth]{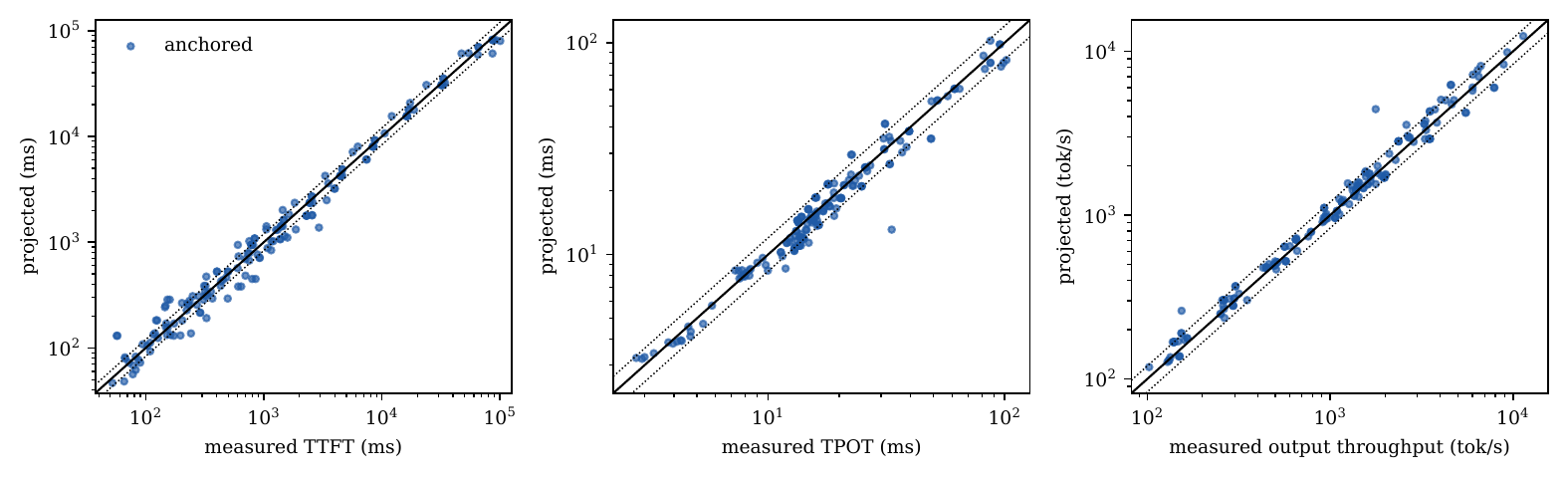}
\caption{Predicted against measured, 172 fixed-length configurations. Dotted lines show $\pm$20\%. Anchored mode stays near the diagonal on all three metrics.}
\label{fig:fixed}
\end{figure*}

\begin{table}[h]
\caption{Fixed-length accuracy on MI355X: median absolute error (\%) per model, anchored mode. Every anchor here is a sweep of 4 to 11 batch sizes. Engines: A=ATOM, S=SGLang, V=vLLM. The output sequence length (OSL) is 1024 for all rows. Tput is output throughput.}
\label{tab:fixed}
\centering\scriptsize\setlength{\tabcolsep}{3pt}
\begin{tabular}{@{}lrlrrr@{}}
\toprule
Model & $n$ & Eng. & TTFT & TPOT & Tput\\
\midrule
MiniMax-M2.7 & 7 & A & 2.7 & 4.5 & 4.7\\
Qwen3-14B-FP8 & 8 & A/S & 4.8 & 1.5 & 1.2\\
GLM-5.2-MXFP4 & 67 & S & 11.7 & 9.5 & 10.6\\
DeepSeek-R1-0528-MXFP4 & 7 & A & 14.9 & 12.1 & 10.9\\
DeepSeek-V4-Flash & 67 & S/V & 15.2 & 6.3 & 7.4\\
Qwen3.6-35B-A3B & 7 & A & 19.1 & 8.1 & 9.0\\
DeepSeek-V4-Pro-MXFP4 & 2 & V & 19.7 & 12.5 & 10.2\\
gpt-oss-120b & 7 & A & 24.7 & 4.0 & 6.4\\
\midrule
ISL 1k / OSL 1k & 99 & & 19.1 & 9.2 & 9.6\\
ISL 8k / OSL 1k & 73 & & 6.3 & 6.4 & 6.9\\
\textbf{All} & \textbf{172} & & \textbf{13.6} & \textbf{8.1} & \textbf{8.8}\\
\bottomrule
\end{tabular}

\end{table}

\begin{table}[h]
\caption{Choice ranking in both modes on the scored corpus of \S\ref{sec:fixed} (42 choice sets; both modes scored on the same 172 configurations and pairs). Pair accuracy is the share of decisive pairs in the right order (ties count half). Top-1 is how often the predicted pick is the measured best. Regret is how much slower the pick is than the best, median/p90 over sets.}
\label{tab:rank}
\centering\scriptsize\setlength{\tabcolsep}{4pt}
\begin{tabular}{@{}ll rrrr@{}}
\toprule
Mode & Metric & Pairs & Pair acc.\ (\%) & Top-1 (\%) & Regret p50/p90 (\%)\\
\midrule
anchored & Throughput & 68 & 93.4 & 73 & 0.0/0.2\\
 & TPOT & 89 & 85.4 & 60 & 0.0/3.9\\
\addlinespace[2pt]
simulate & Throughput & 68 & 89.0 & 59 & 0.0/6.2\\
 & TPOT & 89 & 84.3 & 54 & 0.1/7.0\\
\bottomrule
\end{tabular}

\end{table}

Most remaining ranking errors are pairs that noise or identical execution makes impossible to order. Anchored mode misorders only three throughput pairs. All three are GLM-5.2 SGLang pairs at 8192-token prompts and 256 clients, where repeated runs of the same configuration differ by 25--63\%. On TTFT, 39 of 118 decisive pairs differ only in a flag that does not change execution, so no predictor can order them. On the rest, \sys{} orders 82--84\% correctly.

\begin{table}[h]
\caption{Serving performance models as each paper reports them. Setups differ, so errors are not directly comparable; the table shows what each model needs and where it was validated. The row marked ``run by us'' is our own run, median error on the same rows as \sys{}.}
\label{tab:related}
\centering\scriptsize\setlength{\tabcolsep}{2.5pt}
\begin{tabular}{@{}l>{\raggedright}p{22mm}>{\raggedright\arraybackslash}p{37mm}@{}}
\toprule
System & Step cost from & Reported validation\\
\midrule
Vidur & profiled operators per model and shard; dense only & $<$9\% request latency\\
\quad run by us & profiled on MI355X & 40\% TPOT, 40\% throughput on 30 SGLang rows (\sys{}: 7.2\%, 7.4\%)\\
APEX & profiled operators & 10.7\% on plan speedups, vLLM and SGLang\\
LLMServingSim & accelerator simulators & 14.7\%, vLLM on 4$\times$ RTX 3090\\
AIConfigurator & kernel database per GPU and framework; NVIDIA only & 7.8\% TPOT on one H100 node; 25.5\% throughput on its Pareto picks (TensorRT-LLM)\\
BestServe & roofline, efficiencies tuned per stack; dense only & 8.6--30.1\% over four scenarios\\
LLMCompass & hardware description & 4.1\% inference latency\\
SCOOT & a live run per trial & none needed: every trial is measured\\
\midrule
\sys{} & one step sweep on at most one node per kernel set & 8.1\% TPOT, 8.8\% throughput on 172 configurations, 3 engines; 5.5\% on 159 disaggregated splits; agentic (incl. KV offload) 8\% throughput, 14\% ITL on 95 configurations; MI355X\\
\bottomrule
\end{tabular}
\end{table}

\begin{table}[h]
\caption{Measured Pareto charts against anchored-mode projections. Error is the median relative error per point; $\rho$ is the median within-chart Spearman correlation of throughput per GPU; \emph{frontier hit} counts measured-frontier configurations that are also on the projected frontier. Anchor runs are excluded; the AgentX row is the 95 covered runs.}
\label{tab:pareto}
\centering\scriptsize\setlength{\tabcolsep}{3pt}
\begin{tabular}{lrrrrrr}
\toprule
Corpus & Charts & Points & Tput/GPU err & Interact.\ err & $\rho$ & Frontier hit \\
\midrule
Our fixed-length runs & 11 & 172 & 8.8\% & 8.2\% & 0.99 & 63/80 \\
InferenceX fixed-length & 8 & 262 & 7.0\% & 6.0\% & 0.99 & 65/79 \\
AgentX trace (covered) & 6 & 95 & 4.2\% & 5.2\% & 0.99 & 42/47 \\
\midrule
All & 25 & 529 & 7.0\% & 6.5\% & 0.99 & 170/206 \\
\bottomrule
\end{tabular}

\end{table}

\subsection{Narrower anchors}
\label{app:narrow}

\begin{table}[h]
\caption{TP8 deployments projected from TP4 anchors (fixed-length corpus, anchored mode): median absolute error (\%) and the Pearson correlation $r$ of log projected against log measured throughput. Same-width rows are the corpus rows at their anchor's own TP. ISL is the input sequence length.}
\label{tab:tp4}
\centering\scriptsize\setlength{\tabcolsep}{3pt}
\begin{tabular}{@{}llrrrrr@{}}
\toprule
Model & ISL & $n$ & TTFT & TPOT & Tput & $r$ \\
\midrule
GLM-5.2-MXFP4 (S) & 8k & 50 & 5.7 & 7.7 & 9.2 & \\
GLM-5.2-MXFP4 (S) & 1k & 17 & 22.6 & 15.9 & 19.5 & \\
DeepSeek-R1-0528-MXFP4 (A) & 1k & 7 & 14.9 & 12.1 & 10.9 & \\
Qwen3.6-35B-A3B (A) & 1k & 7 & 19.1 & 8.1 & 9.0 & \\
gpt-oss-120b (A) & 8k & 7 & 24.7 & 4.0 & 6.4 & \\
DeepSeek-V4-Pro-MXFP4 (V) & 1k & 2 & 19.7 & 12.5 & 10.2 & \\
\midrule
TP4$\to$TP8, all & & 90 & 16.5 & 9.3 & 9.7 & 0.994 \\
Same width & & 82 & 11.0 & 5.5 & 6.1 & 0.992 \\
\bottomrule
\end{tabular}
\end{table}

\para{Agentic.} Anchoring three TP8 stacks at TP4 instead projects their ten held-out TP8 runs within 1.08, 1.18 and 1.24, against 1.04, 1.10 and 1.34 from TP8 anchors; the default anchors stay at the target width.

\subsection{Deployment mix on the full trace}
\label{app:mixfull}

\begin{table}[h]
\caption{Deployment mix for six models on MI355X, agentic trace. Per tier: slots where some configuration meets the tier on hardware, picks that are the measured best or within 12\% of its throughput, and picks that miss the tier.}
\label{tab:mix}
\centering\scriptsize
\resizebox{\columnwidth}{!}{\begin{tabular}{@{}lrrrrr@{}}
\toprule
Tier & Interactivity & TTFT & Slots & Best / within 12\% & Miss\\
\midrule
realtime & $\geq$150 tok/s & $\leq$1\,s & 5 & 2 / 3 & 1\\
interactive & $\geq$70 tok/s & $\leq$2.5\,s & 6 & 3 / 4 & 1\\
responsive & $\geq$25 tok/s & $\leq$10\,s & 6 & 4 / 4 & 0\\
bulk & -- & -- & 6 & 5 / 5 & 0\\
\midrule
\textbf{All} & & & \textbf{23} & \textbf{14 / 16} & \textbf{2}\\
\bottomrule
\end{tabular}
}
\end{table}

On the full 184-run trace we restrict the candidates to the 123 runs an anchor covers (same engine build and layout, within $4\times$ of its sessions; the 28 anchors are among them), so that every projection is anchored mode, and score each pick against the measured best of these candidates. All 23 slots keep a candidate that meets the tier. The planner picks that best in 14 and a configuration within \MIXBAND{} of it in 16, with a median loss of 0\%, and its five-candidate short list holds the best in 21 slots and a configuration within \MIXBAND{} in 22.

\subsection{Agent searches}
\label{app:search192}

We ran the agent's seed stage for 24 objectives and four models, each with and without an SLO, on an agentic request shape of 130k prompt and 900 output tokens. This scores 56,328 candidates and uses no LLM. For DeepSeek-V4-Pro, the only model measured where the searches land, the throughput optimum found (TP8, attention DP, FP4, 256 clients) is predicted within 0.3\% of its measured value. Speculative decoding won four of the 24 objectives, all of them per-user latency, and the SLO changed the answer for 8. The searches took 3--8 CPU-hours, and measuring their candidates would take about 25{,}000 GPU-hours (\S\ref{sec:intro}).

\subsection{Disaggregated layouts}
\label{app:disagg}

\begin{table}[h]
\caption{Disaggregated validation, anchored mode: median absolute error (\%) on the 159 InferenceX fixed-length splits on MI355X that we evaluate (four models, 8--24 GPUs, 1--2{,}048 clients). Each pool restores one of 16 anchors built from colocated runs; no split is used as an anchor. TTFT is scored on the mean InferenceX reports.}
\label{tab:splits}
\centering\scriptsize\setlength{\tabcolsep}{3pt}
\begin{tabular}{@{}lrrrr@{}}
\toprule
Runs & $n$ & Tput & Inter. & TTFT \\
\midrule
SGLang + MoRI splits & 58 & 8.5 & 12.6 & 22.2 \\
vLLM + NIXL splits & 37 & 5.2 & 4.0 & 7.1 \\
ATOM + Mooncake splits & 64 & 4.8 & 2.3 & 18.6 \\
\midrule
All splits & 159 & \textbf{5.5} & \textbf{5.3} & \textbf{16.9} \\
\bottomrule
\end{tabular}
\end{table}

\para{TTFT and the KV hand-off.} TTFT needs two things a colocated step model does not have. An anchor's lowest-concurrency TTFT is part GPU prefill and part host time (Hypertext Transfer Protocol (HTTP) handling, tokenization at 5--11\,$\mu$s per prompt token, scheduling); its own decode stalls give the GPU part, so the rest is the host's. A split pays that host time on both servers: at low concurrency the measured DeepSeek-V4-Pro ATOM splits sit 158\,ms above GPU prefill at 8k, about twice the 80\,ms of host time in the colocated anchor. On the prefill server part of it is serial, so \sys{} bills half of the anchor's host time on the prefill step that admits a request, and later prompts queue behind it. This lowers the median split TTFT error from 23\% to 17\%. The first token is stamped when a decode replica admits the request, since the router streams from the decode server. At up to 16 clients the median TTFT error is 2.4\% for vLLM + NIXL, 12\% for SGLang + MoRI and 16\% for ATOM + Mooncake. The KV hand-off is priced as a floor plus the exposed bytes over each prefill GPU's own link. For MoRI these are 47.9\,GB/s and 120\,$\mu$s per batched transfer, from MoRI's remote direct memory access (RDMA) benchmark over the one 400\,Gb/s Pollara NIC per GPU, and only the last layer is exposed, because MoRI streams the cache layer by layer while prefill runs; NIXL and Mooncake send the whole cache after prefill, at 400\,GB/s with a 5\,$\mu$s floor and 200\,GB/s with 10\,$\mu$s. At these prompt lengths the hand-off is small: replacing the 300\,GB/s MoRI rate of an earlier version with the measured 47.9\,GB/s moves no projected TTFT by more than 3\% and leaves every median error unchanged.

\para{Residual rule for agentic splits.} We drop the anchor's end-to-end residual for the 13 agentic splits, since it corrects the colocated scheduler, which a split does not run.

\para{What decides the comparison.} Five modeling details decide whether disaggregation or colocation wins. First, each prefill replica adds capacity. Second, each pool has its own KV budget, summed over its attention-DP ranks. Without this, a DEP8 decode pool holds an eighth of its KV, and the comparison favors colocation. Third, each pool admits at most the requests its engine's recipe allows, and SGLang caps a decode pool without attention DP at 128. Fourth, under attention DP every prefill rank has its own token budget and attends only its own prompts, while the MoE sees all of them. Fifth, an anchor harvested with speculation is not used for a target without it.

\para{On agentic traffic.} We ran the mix planner's rule (\S\ref{sec:mixplanner}) on the AgentX trace in simulate mode, for six models, with colocated and disaggregated layouts as candidates, both swept to 256 clients. It picks a disaggregated layout in three of 19 slots that some layout meets: DeepSeek-V4-Pro at the interactive tier with DEP8 prefill and decode pools, MiniMax-M3 at the interactive tier and Qwen3.5-397B at the realtime tier. InferenceX measured splits of that model on the same trace, and a DEP8 split is also what the planner picks when those measured splits are its candidates, though there in the responsive and bulk tiers (\S\ref{sec:disagg}). The token ratio alone would mislead: at the trace's shape of 130k prompt and 900 output tokens it gives prefill 144 of every 145 GPUs, but prefix reuse leaves about 10{,}400 computed prompt tokens per request, and decode holds each request's full KV cache for all 900 steps, so memory, not decode work, sizes the decode pool. Elsewhere, with a median 97.8\% of prompt tokens hitting the prefix cache, the uncached prefill per turn is short and seldom stalls decode, as with 1k-token prompts. Most other picks combine forms of parallelism: 12 of 19 are TP8 with EP8 and attention DP over all eight ranks.


\section{Harvest checks}
\label{sec:lessons}

An anchor is only as good as the conditions of its harvest, so \sys{} keeps those conditions in the artifact. Each check below stops a measurement from silently describing something other than the deployment. We validated each check on measured runs. None is a fitted correction. The remedy is always to record and check, then re-harvest or refuse.

\para{The prompt length is recorded.} The attention backend decides whether an anchor transfers across prompt lengths. We applied the same 8192-token DeepSeek-V4 anchor at 1024 tokens. On SGLang, whose backend reads a fixed top-$k$ of the cache, it is within 3\%. On vLLM's dense MLA path, it is 41\% high (18.96 vs.\ 13.44\,ms). So anchors keep their decode sweep's prompt length, and sweeps at two lengths give a per-token decode slope.

\para{Weights are real.} An untrained router spreads tokens evenly over experts. With random weights, GLM-5.2 decode is $1.337\times$ too fast at batch 64 and $1.045\times$ at batch 4, and the gap grows with batch, as expert imbalance would cause. Anchors record whether weights were real, and the harvester uses real checkpoints.

\para{Prefix caching is off during harvest.} A length sweep reuses its own prefix by design. With caching on, a DeepSeek-V4-Pro sweep gave 68--89\,ms across 1024--8192 tokens at $R^2=0.995$; with caching off, it gave 143--782\,ms. Nothing downstream can tell the two apart, so the harvester disables caching and the anchor records it.

\para{Width is checked.} A target more than one TP doubling from its anchor is flagged as extrapolated: GLM-5.2 prefill gains $1.80\times$ from TP2 to TP4 but only $1.32\times$ from TP4 to TP8, which a TP4 anchor cannot see. A new engine build needs a new harvest.

\para{Packing probes check themselves.} If the token budget does not admit a probe's sequences together, the probe measures serialization. Two checks catch this: a step of $N$ packed tokens must cost 0.63--1.39$\times$ a single $N$-token sequence (bad probes gave 3.59$\times$ and 7.06$\times$), and the fit must reproduce the probe's widest step. Taking each probe at the billed sequence length gives TTFT errors of $-$2.2, +13.9, $-$4.4 and +1.1\% on four models (Table~\ref{tab:component}).

\para{Mechanisms, not factors.} We model a cost by the mechanism that causes it, and we measure the mechanism. The PP model of \S\ref{sec:pp-model}, for example, uses the token-return hop and the host time SGLang cannot overlap under PP; we measured both, and it has 2.3\% median TPOT error on Qwen3-14B-FP8 with nothing fitted to a PP ladder. Constants in \sys{} are measurements of named quantities, taken once per architecture or engine and applied unchanged everywhere else. The few fitted terms (host time per engine, the node bandwidth $B_\text{node}$, the gpt-oss coverage curve and the share of $I_d$ kept for SGLang) are stated where they are used.

\section{Per-configuration agentic results}
\label{app:var}

Table~\ref{tab:var-rows} breaks the 95 covered agentic configurations of \S\ref{sec:variable} (same layout, image and precision as an anchor, within $4\times$ of its sessions) down by model, engine and precision. The 28 anchors are 7 at TP2, 10 at TP4 and 11 at TP8; 20 run four sessions.

{\small\begin{table}[h]
\caption{Agentic AgentX replay, anchored mode: geometric-mean error factor $\max(r,1/r)$ per model, engine and precision over the covered held-out configurations.}\label{tab:var-rows}
\centering\scriptsize\setlength{\tabcolsep}{4pt}
\begin{tabular}{@{}lllrrrr@{}}
\toprule
Model & Engine & Prec. & $n$ & Tput & ITL & TTFT\\
\midrule
DeepSeek-V4-Pro & ATOM & FP4 & 7 & 1.12 & 1.21 & 1.50\\
DeepSeek-V4-Pro & SGLang & FP4 & 4 & 1.35 & 2.01 & 1.16\\
DeepSeek-V4-Pro & vLLM & FP4 & 5 & 1.07 & 1.14 & 1.65\\
DeepSeek-V4.1-Flash & ATOM & FP4 & 6 & 1.03 & 1.10 & 1.14\\
DeepSeek-V4.1-Flash & SGLang & FP4 & 9 & 1.03 & 1.09 & 1.42\\
DeepSeek-V4.1-Flash & vLLM & FP4 & 8 & 1.05 & 1.04 & 1.08\\
GLM-5.2 & ATOM & FP4 & 9 & 1.04 & 1.11 & 1.07\\
GLM-5.2 & SGLang & FP4 & 7 & 1.06 & 1.13 & 1.35\\
GLM-5.2 & SGLang & FP8 & 6 & 1.06 & 1.13 & 1.47\\
Kimi-K3 & vLLM & FP4 & 5 & 1.04 & 1.05 & 1.20\\
MiniMax-M3 & ATOM & FP4 & 9 & 1.03 & 1.06 & 1.11\\
MiniMax-M3 & vLLM & FP4 & 7 & 1.08 & 1.24 & 1.17\\
Qwen3.5 & SGLang & FP4 & 8 & 1.15 & 1.06 & 1.10\\
Qwen3.5 & SGLang & FP8 & 5 & 1.13 & 1.13 & 1.14\\
\midrule
All covered & & & 95 & 1.08 & 1.14 & 1.23\\
\bottomrule
\end{tabular}
\end{table}
}

\section{Component and collective checks}
\label{app:checks}
\label{sec:components}
\label{sec:comm-eval}

\subsection{Component checks}

\begin{table*}[h]
\caption{Checks of individual modeling decisions against measured MI355X runs.}
\label{tab:component}
\centering\scriptsize\setlength{\tabcolsep}{2.5pt}
\begin{tabular}{@{}p{38mm}p{62mm}p{68mm}@{}}
\toprule
Decision & Setup & Result\\
\midrule
Decode TP transfer (Eq.~\ref{eq:floor}) & DeepSeek-V4-Pro TP4 anchor $\to$ measured TP8, $b$=1--64 & 0.9\% (max 3.5\%) vs.\ 18.5\% (TP4 minus simulated drop), 44.1\% (TP4 halved)\\
Prefill fixed term & DeepSeek-V4-Pro TP4 \& TP8 prefill, solve for $F$ & $F$=146.90 / 145.86\,ms at 4k / 8k-token prompts\\
Anchor width & each TP-ladder rung as anchor for the others & 4 GPUs 6.6\% vs.\ 8 GPUs 7.6\%\\
Fallback width law & Llama-8B / GLM-5.2 TP4$\to$TP8 prefill & serial $s$ plus parallel $p/w$: $-$4.6 / $-$20.2\% vs.\ holding the fitted intercept $-$20.8 / $-$26.5\%\\
Packed prefill at billed length & 4 models probed at 1024-tok sequences & TTFT $-$2.2, +13.9, $-$4.4, +1.1\%\\
Single-wave decode anchor & DeepSeek-V4-Pro TP8 mean TPOT / median ITL & 1.14/1.23/1.41 at $C$=16/32/64: stalls the DES already models\\
PP per-hop cost (Eq.~\ref{eq:pp}) & added TPOT / $p$ at $C$=1, 3 models, PP 2--8 & 920--1130\,$\mu$s per hop; fit $o$=970\,$\mu$s\\
PP host time (Eq.~\ref{eq:pp}) & SGLang overlap off $-$ on, PP1, 3 models, $b$=1--64 & 0.31--0.39\,ms small batch, 0.69--0.78\,ms at 64; $0.33+0.0063\,b$\,ms\\
Irreducible residual & rows equal in every field the projector reads & 2.7\% TTFT, 0.7\% TPOT\\
\bottomrule
\end{tabular}
\end{table*}

Table~\ref{tab:component} checks individual decisions against measured runs. Two need a comment. First, anchors measure a \emph{single wave} of decode. Under continuous batching, every admitted prefill stalls decode; the stalls are rare but large, and at 16--64 concurrent requests they make mean TPOT 1.14--1.41$\times$ median ITL. Feeding a mean-TPOT anchor to a simulator that schedules those stalls itself over-predicts TPOT by about the same amount (+13\%, +23\%, +31\%), so a single-wave anchor avoids counting the scheduler twice. Second, packed-prefill probes must be taken at the sequence length the anchor is billed at, because packing changes with each sequence's length.

\subsection{Collective communication}

\begin{table*}[h]
\caption{Collective model on MI355X against 124 single-node and 35 two-node measurements: constants, mean absolute percentage error (MAPE) per group size, worst point. Custom all-reduce has one $(\alpha, B)$ pair per kernel (one-shot, two-shot); above 8\,MiB in a node, one $B_\text{node}$ is fitted on a single point; across nodes, one bandwidth per collective is fitted on its 1\,GiB point; the datasheet row fits nothing. Points are held out from the fit that predicts them.}
\label{tab:coll}\label{tab:coll-const}
\centering\scriptsize\setlength{\tabcolsep}{5pt}
\begin{tabular}{@{}llllll@{}}
\toprule
Collective (sizes) & GPUs & $\alpha$ ($\mu$s) & $B$ (GB/s) & MAPE (\%) & max (\%)\\
\midrule
\multicolumn{6}{@{}l}{\emph{Inside a node: decode sizes, per-kernel fit, leave-one-size-out}}\\
Custom all-reduce, graph ($\leq$8\,MiB) & 2 / 4 / 8 & 4.9 / 5.5, 6.4 / 6.4, 11.0 & 60.4 / 63, 100 / 56, 123 & 2.5 / 4.1 / 4.5 & 12\\
RCCL all-to-all ($\leq$32\,MiB) & 2 / 4 / 8 & 37.8 / 35.2 / 37.8 & 113.3 / 205.5 / 408.8 & 5.3 / 4.9 / 4.2 & 12\\
\multicolumn{6}{@{}l}{\emph{Inside a node: $>$8\,MiB, one parameter for the whole node}}\\
RCCL all-reduce & 2 / 4 / 8 & 50.8 / 41.1 / 39.0 & mesh, 600 & 1.8 / 1.3 / 5.1 & 13\\
RCCL all-to-all & 2 / 4 / 8 & 40.1 / 36.5 / 39.1 & mesh, 600 & 3.5 / 9.6 / 5.4 & 10\\
RCCL send/recv & 2 & 59.1 & mesh, 600 & 5.8 & 7\\
\multicolumn{6}{@{}l}{\emph{Across two nodes: $>$8\,MiB, one bandwidth per collective}}\\
Send/recv, datasheet NIC & 2 & 44.2 & 40.0 & 9.1 & 13\\
Send/recv & 2 & 44.2 & 45.1 & 3.6 & 11\\
All-reduce & 2 / 16 & 58.7 / 90.5 & 19.8 / 31.6 & 4.7 / 5.3 & 9\\
All-to-all & 16 & 47.4 & 37.9 & 10.1 & 15\\
\bottomrule
\end{tabular}
\end{table*}

\begin{figure*}[h]
\centering
\includegraphics[width=\textwidth]{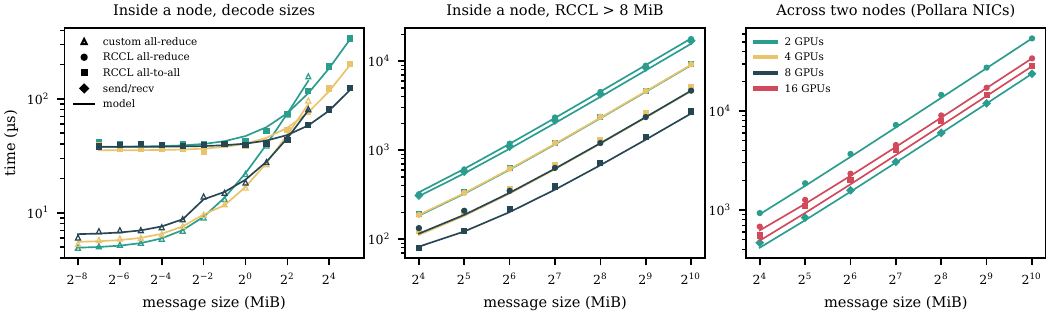}
\caption{MI355X collectives, measured (points) and modeled (lines). Left: decode-size messages on vLLM's custom all-reduce under graph capture and on RCCL all-to-all. Middle: RCCL above 8\,MiB inside a node, where one node parameter predicts every point. Right: RCCL across two nodes over Pollara NICs.}
\label{fig:coll}
\end{figure*}

We measured the collectives a serving step issues on MI355X, on one node and across two. Inside a node: RCCL all-reduce, all-to-all and send/recv from 8\,KiB to 1\,GiB on 2, 4 and 8 GPUs, plus vLLM's custom all-reduce from 4\,KiB to 8\,MiB under graph capture, the way a decode step runs it. Across two nodes: send/recv between stage GPUs, all-reduce with one GPU per node and over all 16 GPUs, and the 16-GPU all-to-all of a two-node EP group, over the eight 400\,Gb/s Pollara NICs of each node on two node pairs. Each point is the median of seven timing windows, timed on the slowest rank. The model tracks every collective, group size and tier to within about 10\% mean error (Figure~\ref{fig:coll}, Table~\ref{tab:coll}).

\para{Decode sizes.} At kilobytes to a few megabytes, the cost comes from the kernel, not the wire. vLLM's custom all-reduce runs a one-shot kernel below its switch size (256\,KiB on eight GPUs, 512\,KiB on four) and a two-shot kernel above it, so \sys{} prices each with its own $\alpha+m/B$. With leave-one-size-out scoring, the fit is within 2.5--4.5\% across 2, 4 and 8 GPUs, and RCCL all-to-all within 4.2--5.3\%. The one-shot floors, 4.9--6.4\,$\mu$s, match the 4.98\,$\mu$s per-kernel occupancy we measured independently from the TP sweep. That is why a step charges only $m/B$ on top of the kernels it already counts.

\para{Large messages: one parameter per node.} Above 8\,MiB, RCCL runs over the xGMI mesh. One number, $B_\text{node}=600$\,GB/s, fitted on a single point (8-GPU, 1\,GiB all-reduce), predicts every other point on every group size: 1.3--5.1\% mean error for all-reduce, 3.5--9.6\% for all-to-all and 5.8\% for send/recv.

\para{Across nodes.} With nothing fitted, the datasheet NIC (50\,GB/s at 0.8 point-to-point efficiency) predicts cross-node send/recv within 9.1\% mean error. This is the hop a pipeline stage boundary uses. Fitting one bandwidth per collective on its own 1\,GiB point and holding out every other size gives 3.6\% for send/recv, 4.7\% (one GPU per node) and 5.3\% (16 GPUs) for all-reduce, and 10.1\% for the two-node all-to-all. Above 1\,MiB, the two node pairs differ by a median of 0.7\%, so one profile per cluster is enough on the pairs we measured. Table~\ref{tab:coll} lists every constant.

\section{Pipeline-parallel measurements}
\label{app:pp}

Table~\ref{tab:pp-abs} gives the absolute measurements behind Table~\ref{tab:pp} for all three models, with the signed anchored-mode error of the per-stage model (Eq.~\ref{eq:pp}). PP1 rows are included for reference; the PP model does not affect them.

{\small\begin{longtable}{@{}lrrrr rrr rrr@{}}
\caption{Pipeline-parallel ladders on SGLang: measured values and signed anchored-mode error (\%) of the per-stage model. TTFT and TPOT in ms, throughput in tok/s.}\label{tab:pp-abs}\\
\toprule
 & & & & & \multicolumn{3}{c}{Measured} & \multicolumn{3}{c}{Error}\\
Model & ISL & TP & PP & $C$ & TTFT & TPOT & Tput & TTFT & TPOT & Tput\\
\midrule
\endhead
Llama-3.1-8B & 1k & 1 & 1 & 1 & 29 & 4.46 & 223 & -7 & -3 & +4\\
Llama-3.1-8B & 1k & 1 & 1 & 16 & 185 & 5.03 & 3071 & +32 & +13 & -12\\
Llama-3.1-8B & 1k & 1 & 1 & 64 & 514 & 7.73 & 7773 & +19 & +21 & -17\\
Llama-3.1-8B & 1k & 1 & 2 & 1 & 35 & 6.32 & 157 & -5 & -1 & +1\\
Llama-3.1-8B & 1k & 1 & 2 & 16 & 170 & 7.40 & 2117 & -10 & +7 & -6\\
Llama-3.1-8B & 1k & 1 & 2 & 64 & 375 & 8.79 & 6988 & +15 & +20 & -17\\
Llama-3.1-8B & 1k & 2 & 2 & 1 & 32 & 5.75 & 173 & -2 & -10 & +11\\
Llama-3.1-8B & 1k & 2 & 2 & 16 & 238 & 6.65 & 2325 & -9 & -3 & +3\\
Llama-3.1-8B & 1k & 2 & 2 & 64 & 368 & 7.76 & 7882 & +75 & +13 & -14\\
Llama-3.1-8B & 1k & 1 & 4 & 1 & 40 & 8.15 & 122 & +0 & +1 & -1\\
Llama-3.1-8B & 1k & 1 & 4 & 16 & 144 & 9.78 & 1614 & -31 & +0 & +0\\
Llama-3.1-8B & 1k & 1 & 4 & 64 & 320 & 10.71 & 5809 & +3 & +5 & -5\\
Llama-3.1-8B & 1k & 1 & 8 & 1 & 52 & 12.06 & 83 & +2 & +0 & -0\\
Llama-3.1-8B & 1k & 1 & 8 & 16 & 112 & 15.06 & 1056 & -29 & -1 & +1\\
Llama-3.1-8B & 1k & 1 & 8 & 64 & 217 & 16.50 & 3833 & +3 & -3 & +3\\
Llama-3.1-8B & 8k & 1 & 1 & 1 & 144 & 4.57 & 213 & +5 & -3 & +2\\
Llama-3.1-8B & 8k & 1 & 1 & 16 & 1146 & 8.19 & 1720 & +14 & +9 & -9\\
Llama-3.1-8B & 8k & 1 & 1 & 64 & 3949 & 21.56 & 2519 & +21 & +4 & -6\\
Llama-3.1-8B & 8k & 1 & 2 & 1 & 182 & 7.18 & 136 & -14 & -11 & +12\\
Llama-3.1-8B & 8k & 1 & 2 & 16 & 1008 & 9.42 & 1539 & -26 & +1 & +2\\
Llama-3.1-8B & 8k & 1 & 2 & 64 & 2891 & 17.44 & 3160 & -12 & -2 & +3\\
Llama-3.1-8B & 8k & 2 & 2 & 1 & 157 & 6.73 & 145 & -8 & -22 & +28\\
Llama-3.1-8B & 8k & 2 & 2 & 16 & 910 & 8.04 & 1792 & -22 & -8 & +10\\
Llama-3.1-8B & 8k & 2 & 2 & 64 & 2402 & 13.08 & 4151 & +0 & -1 & +1\\
Llama-3.1-8B & 8k & 1 & 4 & 1 & 192 & 9.18 & 107 & -15 & -9 & +10\\
Llama-3.1-8B & 8k & 1 & 4 & 16 & 699 & 11.56 & 1308 & -34 & -9 & +12\\
Llama-3.1-8B & 8k & 1 & 4 & 64 & 1827 & 16.54 & 3494 & -24 & -12 & +15\\
Llama-3.1-8B & 8k & 1 & 8 & 1 & 268 & 14.24 & 69 & -35 & -14 & +17\\
Llama-3.1-8B & 8k & 1 & 8 & 16 & 553 & 17.05 & 910 & -40 & -10 & +13\\
Llama-3.1-8B & 8k & 1 & 8 & 64 & 1227 & 19.12 & 3151 & -33 & -8 & +10\\
\midrule
Qwen3-14B-FP8 & 1k & 1 & 1 & 1 & 44 & 8.86 & 112 & +9 & -0 & +0\\
Qwen3-14B-FP8 & 1k & 1 & 1 & 16 & 353 & 10.26 & 1510 & +56 & -0 & -1\\
Qwen3-14B-FP8 & 1k & 1 & 1 & 64 & 1121 & 12.79 & 4610 & +22 & +2 & -4\\
Qwen3-14B-FP8 & 1k & 1 & 2 & 1 & 46 & 11.12 & 90 & +23 & -3 & +3\\
Qwen3-14B-FP8 & 1k & 1 & 2 & 16 & 344 & 12.74 & 1224 & -7 & -2 & +2\\
Qwen3-14B-FP8 & 1k & 1 & 2 & 64 & 812 & 14.65 & 4147 & +15 & -1 & +0\\
Qwen3-14B-FP8 & 1k & 1 & 4 & 1 & 50 & 13.02 & 77 & +30 & -2 & +2\\
Qwen3-14B-FP8 & 1k & 1 & 4 & 16 & 269 & 15.03 & 1047 & -29 & -3 & +3\\
Qwen3-14B-FP8 & 1k & 1 & 4 & 64 & 659 & 16.26 & 3789 & +5 & -3 & +2\\
Qwen3-14B-FP8 & 1k & 1 & 8 & 1 & 55 & 16.93 & 59 & +48 & -2 & +2\\
Qwen3-14B-FP8 & 1k & 1 & 8 & 16 & 175 & 20.03 & 793 & -23 & -2 & +2\\
Qwen3-14B-FP8 & 1k & 1 & 8 & 64 & 569 & 21.48 & 2906 & -23 & -4 & +5\\
\midrule
Qwen3.6-35B-A3B & 1k & 1 & 1 & 1 & 53 & 3.95 & 250 & -70 & +0 & +1\\
Qwen3.6-35B-A3B & 1k & 1 & 1 & 16 & 259 & 7.51 & 2063 & -70 & +2 & +1\\
Qwen3.6-35B-A3B & 1k & 1 & 1 & 64 & 504 & 12.82 & 4811 & -62 & -2 & +5\\
Qwen3.6-35B-A3B & 1k & 1 & 2 & 1 & 58 & 6.19 & 160 & -65 & -5 & +6\\
Qwen3.6-35B-A3B & 1k & 1 & 2 & 16 & 264 & 9.52 & 1638 & -83 & -9 & +13\\
Qwen3.6-35B-A3B & 1k & 1 & 2 & 64 & 344 & 13.25 & 4713 & -65 & -3 & +5\\
Qwen3.6-35B-A3B & 1k & 1 & 4 & 1 & 64 & 8.17 & 122 & -63 & -4 & +5\\
Qwen3.6-35B-A3B & 1k & 1 & 4 & 16 & 319 & 11.33 & 1375 & -90 & -12 & +17\\
Qwen3.6-35B-A3B & 1k & 1 & 4 & 64 & 265 & 14.88 & 4229 & -68 & -11 & +14\\
Qwen3.6-35B-A3B & 1k & 1 & 8 & 1 & 82 & 12.43 & 80 & -61 & -6 & +6\\
Qwen3.6-35B-A3B & 1k & 1 & 8 & 16 & 324 & 16.44 & 955 & -89 & -10 & +13\\
Qwen3.6-35B-A3B & 1k & 1 & 8 & 64 & 200 & 18.90 & 3355 & -71 & -11 & +14\\
\midrule
\bottomrule
\end{longtable}
}

\section{Anchor artifact}
\label{app:anchor}

An anchor is one JSON file per (model, engine, anchor-defining choices, width). Its fields:
\begin{itemize}
\item \textbf{Signature:} accelerator architecture, engine and image, model checkpoint, weight/activation/KV/expert dtypes, attention and MoE backend, graph-capture mode and sizes, speculative method, TP/EP/attention-DP width.
\item \textbf{Decode sweep:} median ITL of a single wave at each captured batch size, at a recorded prompt length. Optional sweeps at a second prompt length enable context differencing.
\item \textbf{Prefill sweep:} TTFT at concurrency 1 for five prompt lengths with prefix caching off, and the fitted $F$, $c_1$, $c_2$.
\item \textbf{Packed-prefill probe} (optional): step time with $k$ sequences of the billed length. The token budget is recorded so serialization can be detected.
\item \textbf{Provenance:} harvest timestamp, node, whether weights were real, whether prefix caching was off, and the client command line.
\end{itemize}
Anchors live in a cache directory, and projection selects them by signature; a benchmark is rerun only for a kernel set or width the cache does not hold. Projection refuses an anchor whose accelerator or engine differs from the target. If its prompt length, width or packing length differs, projection uses it but flags it.

\section{Tuning agent details}
\label{app:agent}

\begin{figure}[h]
\centering
\fbox{\begin{minipage}{0.95\columnwidth}\scriptsize
\textbf{Input:} model card, cluster, objective, SLO budgets, request profile, lever list.\\
\textbf{Ledger:} every trial with its configuration, metrics, legality and reason.
\begin{enumerate}
\item \textbf{Seed.} Build the baseline. Emit the ordered seed list (TP, attention DP, precision, load, scheduler, speculation, EP and MoE options, collectives, disaggregation). Keep the first $N_\text{seed}$ legal, distinct seeds.
\item \textbf{Score.} For each candidate: check structure; project it with the anchor of its parallel shape; reject it if memory exceeds the HBM cap or a latency budget is missed; log it.
\item \textbf{Plan.} The planner LLM reads the ledger and emits hypotheses and 5--8 partial configurations as JSON.
\item \textbf{Investigate.} A recursive LM runs the plan with tools: evaluate, read history, read best, read budget, write notes, ask a sub-question. Each evaluation goes through step 2.
\item \textbf{Stop} when the budget is used, or when the best trial has not changed after 70\% of the budget. Return the best legal trial.
\end{enumerate}
\textbf{Mix planner:} for each model and tier, project every candidate, keep those that meet the tier with margin, and select one (\S\ref{sec:mixplanner}).
\end{minipage}}
\caption{The \sys{} tuning agent. Each candidate costs one projection of under half a second, and no GPU.}
\label{fig:agent}
\end{figure}

\para{Fusions.} Fusions enter the search like any other lever, and every trial reports which fusions would pay off. The legality check has three rules. It offers the quant-fed patterns only with low-precision activations and the MoE patterns only on MoE models, and it rejects a list that repeats a fusion or combines named fusions with the generic flag. The seed planner adds the full legal set and its BF16-safe subset. Every trial reports its \emph{fusion opportunities}, the legal fusions worth at least 2\% of that trial's decode step, largest first. The LLM loop reads them from the ledger and proposes fusions together with the configuration they pay on. The agent thus returns a deployment together with the kernel work most worth doing for it.

\begin{table}[h]
\caption{Named fusions the agent proposes. Each is priced from the step the predictor already charges.}
\label{tab:fusion}
\centering\scriptsize\setlength{\tabcolsep}{3pt}
\begin{tabular}{@{}lp{0.70\columnwidth}@{}}
\toprule
Fusion & Replaces \\
\midrule
\texttt{norm\_quant} & RMSNorm and the activation quant feeding the attention and multilayer perceptron (MLP) GEMMs \\
\texttt{add\_rmsnorm} & residual add and the following RMSNorm, twice per layer \\
\texttt{moe\_expert} & expert gate/up, activation, quant and down in one kernel \\
\texttt{moe\_routing} & router top-$k$ and permute \\
\texttt{act\_quant} & sigmoid linear unit (SiLU)$\cdot$mul and quant feeding the down projection \\
\texttt{rope\_kvstore} & rotary position embedding (RoPE) into the KV-cache store \\
\bottomrule
\end{tabular}
\end{table}

\para{Backbone.} The planner and the investigation loop are DSPy modules over one chat model, reached through LiteLLM, so any provider can be substituted without changing the agent; response caching is off. The 192 searches of \S\ref{sec:search} ran only the deterministic seed stage over a shared grid of candidates, with no LLM call, so their results are reproducible and do not depend on a backbone.

\para{What the LLM sees.} Both stages receive the same five inputs as JSON or compact text: the architecture record (layers, hidden size, heads, MoE and sparse-attention fields), the target cluster (nodes, GPUs per node, HBM), the legal value set of every lever, the objective and its direction, and a summary of the trials so far (the 40 most recent, \S\ref{sec:search-design}). The operator may add free-form guidance, and a scratchpad carries notes across rounds.

\para{Planner prompt.} The planner is asked to design a search before using any tool: which levers are most likely to move the objective for this model, cluster and request profile, in what order to try them, and what to do if they fail. The prompt names the trade-offs to account for: latency against throughput (batch, concurrency, request rate), KV capacity (KV dtype, memory fraction, block size), prefill cost (chunk size, token budget), MoE communication (EP, DeepEP, load balancing), disaggregation, and the kernel options. It returns a short rationale, at most five testable hypotheses, 5--8 partial configurations as a JSON array ordered from most to least promising, and a polish plan. A partial configuration names only the levers it changes; the rest come from the baseline.

\para{Investigation prompt and tools.} The search stage receives the plan and runs a loop. In each iteration it reads the scratchpad and the history, picks a configuration not yet tried, evaluates it, and every 3--5 evaluations writes what it learned. It stops when the budget is spent or the incumbent stops improving. Its tools are: evaluate a configuration (one projection, returning metrics, a signed score and, if the configuration is illegal, the reason); read the history; read the incumbent; read the legal sets, architecture, cluster and objective; read the remaining budget; write or read the scratchpad; and ask a sub-question of the same LLM. It returns the best configuration, a summary of what worked, and suggested follow-ups.

\para{Guards.} The LLM never prices a configuration. Every proposal goes through the same structural check, memory cap, SLO check and projection as the seed sweep's, and a value outside a lever's legal set is rejected with its reason. Sparse attention is offered only at the width the checkpoint was trained with, and, when the operator lists validated weight precisions, only those are offered. The 192 searches scored 56,328 candidates, about 293 per search, including the seeds.

\section{Reproduction}
\label{app:repro}

\para{Software.} Engines ran from public container images: ATOM \texttt{rocm/atom-dev:nightly\_202607231538} and \texttt{nightly\_202608280858}, SGLang \texttt{lmsysorg/sglang-rocm:v0.5.19-rocm720-mi35x} (the September 2026 dated tags), and vLLM \texttt{vllm/vllm-openai-rocm:v0.30.0} plus vLLM built from its public source at commit \texttt{29468dde}, whose nightly image is no longer hosted. All are on Docker Hub. Agentic replays used the public images recorded with each InferenceX run (\texttt{rocm/atom-dev}, \texttt{lmsysorg/sglang-rocm} and \texttt{vllm/vllm-openai-rocm} nightlies). Inside a node, collectives were measured with PyTorch's RCCL process group and vLLM's custom all-reduce in the vLLM image. Across nodes, they were measured in the SGLang image over RDMA over Converged Ethernet (RoCE) on the Pollara NICs, with GPU-direct RDMA through DMA-BUF, rings pinned to one rail per NIC, and cross-rail traffic relayed through a peer GPU on the same node (RCCL's PXN mode). Fixed-length clients used the vLLM \texttt{benchmark\_serving} random dataset with \texttt{-{}-ignore-eos} and $10C$ prompts per point. The agentic replay used AIPerf.

\para{Artifacts.} One script produces each number in the paper from one result file. \texttt{score\_corpus.py} scores the fixed-length corpus in both modes. \texttt{comm\_bench.py} (one node and two) and \texttt{comm\_correlate2.py} produce the collective data. \texttt{pp\_bench.sbatch}, \texttt{pp2n.sbatch} and \texttt{pp\_project.py} produce the pipeline-parallel study on one node and across two, and \texttt{overlap\_probe.sbatch} the host-time probe of Eq.~\ref{eq:pp}. \texttt{rank\_choice.py} produces the choice-ranking scores and \texttt{time\_proj.py} the projection timings. The Pareto sweep and its comparison with measured runs come from the projector's \texttt{pareto} command. We will release the anchors, the measured corpora and these scripts with the camera-ready version.

\para{Search study.} The 24 objectives are: throughput (total, decode and prefill, each per GPU and for the fleet); latency and interactivity (TTFT, TPOT, end-to-end latency, the uncontended decode step, per-user interactivity, per-request generation rate); capacity (maximum concurrent sequences, maximum sustainable concurrency); memory (per GPU, KV cache, weights, activations); prefill/decode interference (mixed-step fraction, TPOT pollution, mixed decode step); communication time (prefill, decode); and the GPU cost of a replica. Each objective of \S\ref{sec:search} is searched with and without the SLO bound. Each engine's legality predicate removes the levers it does not support.

\amdattribution

\end{document}